\documentclass[preprint,10pt,authoryear]{elsarticle}

\usepackage{graphicx}
\usepackage{amsmath}
\usepackage{algorithm}
\usepackage{algpseudocode}
\usepackage{array}
\usepackage{amssymb}
\usepackage{multirow}
\usepackage{float}
\usepackage{longtable}
\usepackage{soul}
\usepackage{hyperref}
\usepackage{cleveref}
\usepackage{graphicx}
\usepackage{caption}
\usepackage{subcaption}

\usepackage{adjustbox}
\usepackage[table,xcdraw]{xcolor}
\usepackage[graphicx]{realboxes}
\usepackage{url}

\journal{Transportation Research Part C} 

\begin{document}
\begin{frontmatter}

\title{Machine Learning-Enhanced Aircraft Landing Scheduling under Uncertainties}

\affiliation[inst1]{organization={School for Engineering of Matter, Transport and Energy},
addressline={Arizona State University}, 
city={Tempe},
postcode={85287}, 
state={AZ},
country={USA}}

\author[inst1]{Yutian Pang}
\author[inst1]{Peng Zhao}
\author[inst1]{Jueming Hu}
\author[inst1]{Yongming Liu\corref{mycorrespondingauthor}}
\cortext[mycorrespondingauthor]{Corresponding author.}
\ead{Yongming.liu@asu.edu}

\begin{highlights}
\item We identify the problem of interest by investigating the various flight scenarios and observe that holding patterns exist in most of the arrival delay cases.
\item We propose using machine learning to predict the estimated arrival time (ETA) distributions for landing aircraft from real-world flight recordings, which are further used to obtain the probabilistic Minimal Separation Time (MST) between successive arrival flights.
\item We propose to incorporate the predicted MSTs into the constraints of the Time-Constrained Traveling Salesman Problem (TSP) for Aircraft Landing Scheduling (ALS) optimization. 
\item We build a multi-stage conditional prediction algorithm based on the looping events, and find that explicitly including flight event counts and airspace complexity measures can benefit the model prediction capability. 
\item We demonstrate that the proposed method reduces the total landing time with a controlled reliability level compared with the First-Come-First-Served (FCFS) rule by running experiments with real-world data.
\end{highlights}

\begin{abstract}
Aircraft delays lead to safety concerns and financial losses, which can propagate for several hours during extreme scenarios. Developing an efficient landing scheduling method is one of the effective approaches to reducing flight delays and safety concerns. Existing scheduling practices are mostly done by air traffic controllers (ATC) with heuristic rules. This paper proposes a novel machine learning-enhanced methodology for aircraft landing scheduling.  Data-driven machine learning (ML) models are proposed to enhance automation and safety. ML enhancement is adopted for both prediction and optimization. First, the flight arrival delay scenarios are analyzed to identify the delay-related factors, where strong multimodal distributions and arrival flight time duration clusters are observed. A multi-stage conditional ML predictor is proposed for improved prediction performance of separation time conditioned on flight events. Next, we propose incorporating the ML predictions as safety constraints of the time-constrained traveling salesman problem formulation. The scheduling problem is then solved with mixed-integer linear programming (MILP). Additionally, uncertainties between successive flights from historical flight recordings and model predictions are included to ensure reliability. We demonstrate the real-world applicability of our method using the flight track and event data from the Sherlock database of the Atlanta Air Route Traffic Control Center (ARTCC ZTL). The case studies provide evidence that the proposed method is capable of reducing the total landing time by an average of 17.2\% across three case studies, when compared to the First-Come-First-Served (FCFS) rule. Unlike the deterministic heuristic FCFS rule, the proposed methodology also considers the uncertainties between aircraft and ensures confidence in the scheduling. Finally, several concluding remarks and future research directions are given. The code used can be retrieved from \href{https://github.com/ymlasu/para-atm-collection/tree/master/flight-operations}{[$\mathsf{Link}$]}.
\end{abstract}


\end{frontmatter}

\section{Introduction\label{sec: introduction}}
The civil aviation industry is losing air traffic control talents, while the need for maintaining daily operations keeps surging \citep{faareport}. This situation leads to increased operational costs, higher safety concerns, an elevated workload for air traffic controllers, and frequent flight delays \citep{dhief2020predicting}. Flight delay is a major problem of interest faced by domain experts, which results in both economic and customer loyalty losses \citep{vlachos2014drivers}. It's reported that $20\%$ of the civil flights in the U.S. were delayed from 2010 to 2018, and the annual cost of delays before the pandemic is estimated to be $\$30$ billion \citep{faa2019}. The initial flight delays come from various resources (e.g., extreme weather conditions, carrier and controller issues) and can propagate through several hours \citep{jetzki2009propagation, kafle2016modeling}. Moreover, the aviation industry is encountering a shortage of experienced operation talents after the COVID-19 pandemic due to various reasons (e.g., loss of operational and airline experience, staffing, and changing customer demand patterns). All of this urges the automation and digitization of the aviation industry in a regulated fashion, which heavily relies on innovative data-driven modeling techniques.

Automated computer-aided decision support tools (DSTs) are practical solutions to address safety and efficiency concerns (e.g., flight delays), with the help of modernized data monitoring and recording equipment. DSTs will help maximize the operational capacity of the terminal maneuvering area (TMA), where the optimization of departure/arrival operations in the TMA is a critical problem of air traffic control (ATC). In most cases, the heuristic decision by the ATC will be suggested together with a graphic view of each corresponding location and speed of the aircraft near the TMA. This setup is efficient on normal operations but leads to flight delays and elevated controller workload during extreme scenarios. The unfolded diamond shape symbols in the graphic view may overlap and lead to significant delays during certain extreme cases \citep{loft2007modeling}. DSTs are developed to alleviate flight delays and maximize operational capacity during certain cases and busy traffic. For instance, the measurement coverage of NextGen will be enlarged to hundreds of nautical miles (NM) due to the advanced surveillance radar for the Automatic Dependent Surveillance-Broadcast (ADS-B) system \citep{adsb}. The enlarged surveillance measurement space enables the possibility of developing optimization-based DSTs, to be applied in the en-route phase. Lastly, DSTs assist controllers in suggesting reasonable resolutions by searching from historical data or learning from human preferences. Various government agencies proposed advanced DST system concepts. \textit{Airport Collaborative Decision Making} (A-CDM) \citep{bolic2021sesar} concept and \textit{Next Generation Air Transportation System} (NextGen) \citep{faa2013nextgen} was proposed by the European Organization for the Safety of Air Navigation (EUROCONTROL) and Federal Aviation Administration (FAA) to assist air traffic controllers in decision makings, with enhanced safety, efficiency, and capacity. Field demonstrations on either single-airport or multi-airport scenarios show great safety enhancements and efficiency improvements \citep{huet2016cdm,ging2018airspace}.


While the government-led efforts mostly focus on building the system workflow for onboard deployment, academic research focuses on algorithmic development and advanced data analytics to enable automated decision-making to support aviation digitization. The Aircraft Landing Scheduling (ALS) problem is vital to overcome flight delays and achieve efficient aviation operations in the TMA \citep{sama2014optimal}. ALS studies the planning of the landing schedule for all the aircraft landed on the same runway in a short time period \citep{yu2011real}, where the runway capacity is pre-defined based on the existing infrastructure \citep{bennell2017dynamic}. In aviation, the ALS problem is viewed as a critical element of the general planning system of aircraft around the TMA \citep{sama2014optimal}. Researchers who are studying ALS focus on the following objectives, 

\begin{itemize}
    \item Maximize the fuel efficiency by arranging the landing aircraft at the most economic landing times and speed profiles \citep{ernst1999heuristic,beasley2000scheduling,pinol2006scatter}.
    \item Minimize the difference to the flight schedules \citep{beasley2001scheduling, bianco2006scheduling,zufferey2022local}.
    \item Maximize the runway throughput by minimizing the total landing time \citep{bianco1999minimizing,atkin2007hybrid,wided2022effective}.  
\end{itemize}

This paper focuses on the last item, i.e., maximizing the runway throughput. During arrivals, the air traffic controllers (ATCs) give instructions to the pilots when the aircraft enters the range of the terminal surveillance radar. Thus, ATCs provide guidance for safe and effective landings. Landing safety is enforced by the Minimum Separation Time (MST) between two landing aircraft. The MST is introduced to account for aerodynamic safety considerations \citep{bennell2017dynamic}. For instance, when the leading aircraft is much heavier than the following aircraft, the leading aircraft's wake vortices will result in hazardous conditions for the following lighter aircraft within MST and poses immediate safety concerns. The ALS problem has been formulated into two sub-problems \citep{yu2011real}. Firstly, the order of the aircraft entering the TMA is determined. Then, the exact scheduled landing time is determined based on the landing sequence and MST. These two steps can be collaboratively solved with proper optimization algorithms. The extension of the surveillance area enables the possibility of developing a novel landing scheduling scheme that can be performed in the en-route phase rather than only in the terminal area to prevent congestion and reduce congestion-related safety concerns. However, the current literature either focuses on formulating the optimization problem in both static \citep{beasley2000scheduling} and dynamic \citep{beasley2004displacement,bennell2017dynamic} scenarios with synthetic examples, or considering one of the related factors during sequencing to formulate the mathematical model (e.g., ground staff workload \citep{boysen2011scheduling}, airline preferences \citep{soomer2008scheduling}). The above pure simulated demonstration limits the applicability and generalizability of the developed algorithms to be deployed in the real world.




\begin{figure}[H]
    \centering
    \begin{subfigure}[t]{0.95\textwidth}
        \centering
        \includegraphics[width=\textwidth]{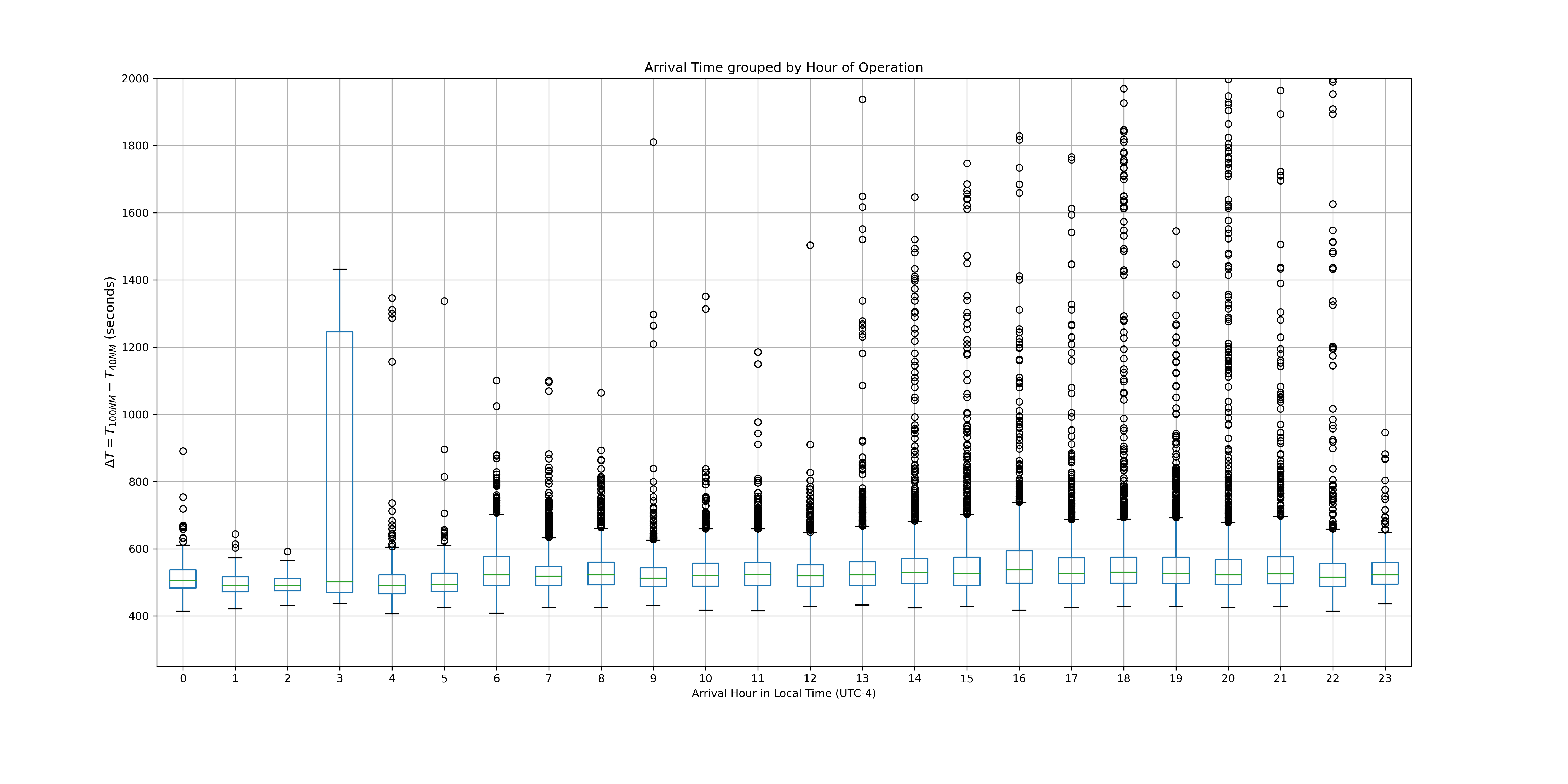}
        \caption{Boxplot grouped by daily hours. It's obvious that the busy hour typically starts from 13:00 to 22:00 each day.}
        \label{fig:boxplot hours}
    \end{subfigure}
    \\ 
    \begin{subfigure}[t]{0.95\textwidth}
        \centering
        \includegraphics[width=\textwidth]{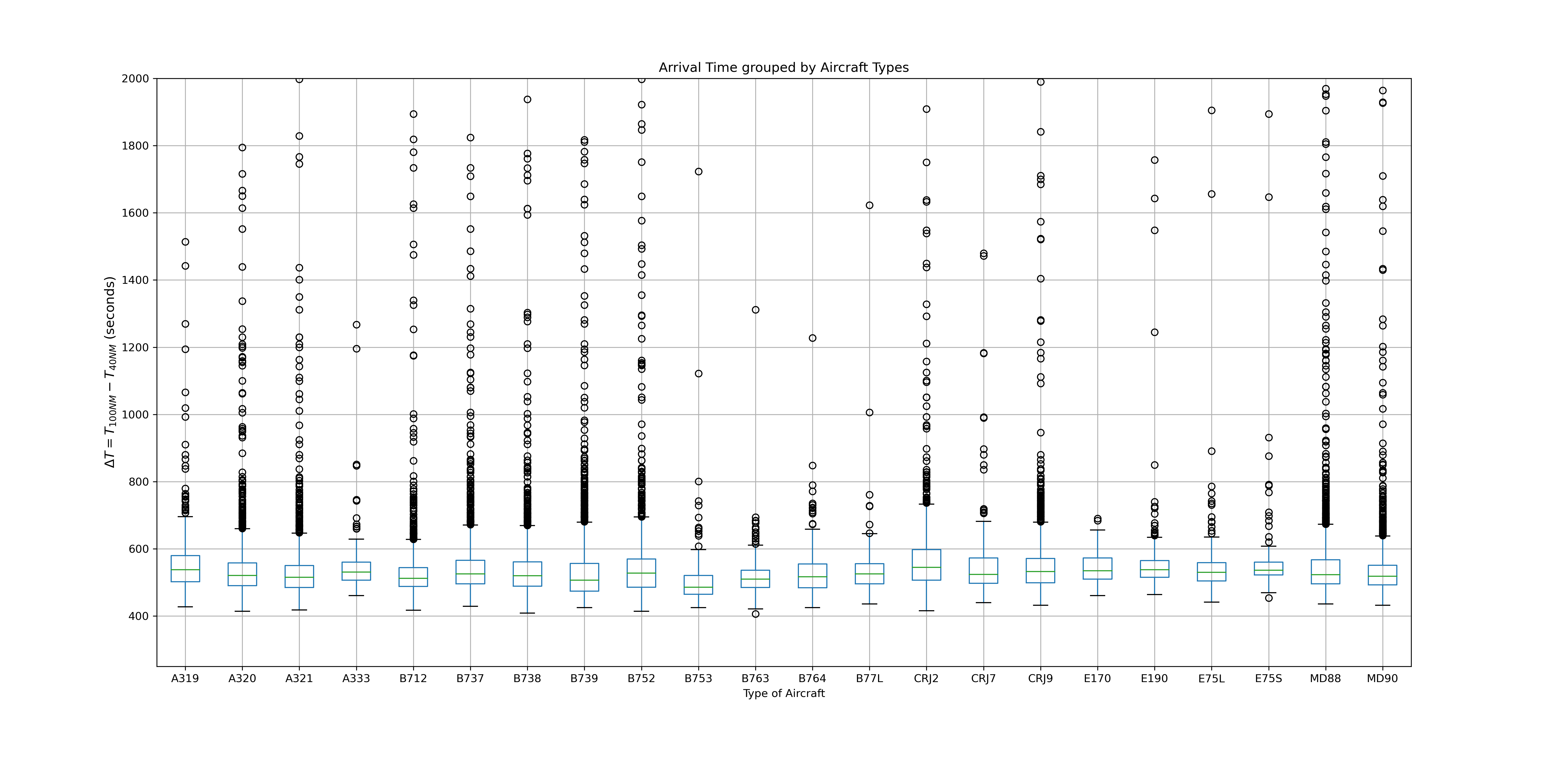}
        \caption{Boxplot grouped by aircraft types. There is not a clear correlation between time spent in TMA to aircraft types.}
        \label{fig:boxplot actype}
    \end{subfigure}
    \caption{Time spent from 100NM to 40NM for landing aircraft entering the TMA of KATL during the entire month of August 2019.}
    \label{fig: boxplots}
\end{figure}

The availability of a well-maintained aviation data warehouse \citep{arneson2019sherlock, jones1996sources} has enabled the possibility of learning and generating aviation operational decisions from realistic operational data. Machine Learning (ML) is an example of data analytics that draws interest from both academia and industry. In the ATM domain, the use of ML techniques also surges in recent years \citep{xu2021multi,zeng2022aircraft,pang2019recurrent,pang2021data, pang2022bayesian,tran2022aircraft}, although multiple challenges (i.e., data privacy/collection/storage/integrity, system reliability, and scalability) still exist when deploying ML systems into real-world (MLOps) \citep{sculley2015hidden,zhu2018big}. Compared to the conventional methods, ML methods show the following benefits: a) A ML-based DST takes advantage of realistic historical data to simulate the human experience accumulation process, where the model can provide \textit{experienced} guidance within the machine response time; b) ML methods are highly flexible to fuse structured or unstructured data from various sources for decision-making. Nonetheless, criticisms against ML methods also rise regarding model interpretability/explainability, prediction generalizability, and output trustworthiness. The authors believed that ML-based DSTs are beneficial for computer-assisted decision-making under the supervision of human controllers. 

In view of the above discussion, there is a need and gap to develop data-driven aircraft landing scheduling algorithms from extended airspace to maximize runway throughput and reduce flight delays. In this paper, we first investigate and identify several factors causing flight delays through data analysis. Then, we propose a data-enhanced optimization technique for ALS, where the statistics of MST is incorporated into the safety-critical constraints under the Traveling Salesman Problem (TSP) formulation. The probabilistic MST is learned with a conditional tree-based ML method, namely a conditional gradient boosting machine (conditional GBM) with quantile distributions to retrieve the upper and lower bounds of MST. Following this, an optimal method using TSP formulation solved by MILP is proposed for sequencing to minimize total delay while taking into account the uncertainty of arrival time prediction. 

The contributions of this work can be summarized as
\begin{itemize}
    \item We investigate several arrival delay scenarios that occurred in the historical data and gain the following insights, a) the arrival time of aircraft has a highly multimodal distribution conditioned on the flight events; b) go around (looping) - the event happened in most arrival delay scenarios - occurs at approximately 100 nautical miles away from the terminal, where FCFS rule starts to take effect; c) the preference of landing scheduling made by human controllers may not be optimal (e.g., landing aircraft from west of terminal should yield to other directions on a west heading runway); d) including weather features can further improve the landing time prediction accuracy. These observations give insights into identifying relevant impact factors in building ALS solutions.
    \item The statistics of MSTs are predicted with a tree-based probabilistic machine-learning algorithm from the historical flight recordings. The obtained probabilistic MSTs are incorporated as safety constraints to the time-constrained traveling salesman problem. To the author's best knowledge, this type of probabilistic scheduling setting is the first time.
    \item We propose to use a conditional ML predictor based on the event counts within a certain distance of the target aircraft to improve the prediction performance. Geographical location, speed profiles, flight event counts, weather features, and airspace complexity measures are integrated together for probabilistic prediction of arrival time, which has not been explored in the open literature.
    \item The proposed framework shows a reduction in total aircraft landing time compared to the FCFS rule, through case studies during busy operation hours at KATL. The proposed method takes effect from extended airspace (e.g., en-route phase flights 200NM away from the terminal), such that early adjustment of aircraft speed profiles can be issued to avoid holding patterns.
\end{itemize}

The rest of the paper is organized as follows. In \Cref{sec: literature review}, we review the related literature on this topic first. The methodology proposed in this research is discussed in \Cref{sec: methodologies}, where the optimization formulation and machine learning predictor are discussed. Investigations on flight delay scenarios and insights are discussed in \Cref{sec: investigations}. The optimization case studies and experimental results are shown in \Cref{sec: experiments}. Finally, conclusions and future insights are given based on the current investigations.

\section{Literature Review\label{sec: literature review}}
This section discusses the related literature to our proposed study on data-enhanced ALS. We first review the studies for the prediction of aircraft estimated arrival time (ETA) and MST in \Cref{subsec: review-prediction}. Then, we review the research on aircraft landing scheduling problems in \Cref{subsec: review-als}.

\subsection{Estimated Arrival Time Prediction and Minimum Separation Time (MST)\label{subsec: review-prediction}}
Landing aircraft move along the predefined landing procedures with standard descending profiles when entering the TMA, with the help of necessary guidance from ATCs. The MST between two consecutive landing aircraft should be guaranteed in the approaching phase. The MST depends on the types and relative positions of two consecutive landing aircraft, which can be translated by considering the speed profiles \citep{sama2014optimal}. Once a landing aircraft enters TMA, it should line up and proceed to the runway. However, delays happen on a daily basis and can propagate from ground to mid-air airplanes due to the sub-optimal scheduling of runway usage. In this case, ATC issues a \textit{holding} order to the approaching aircraft and forces the aircraft to circle around and wait for the clearance to land. The conservative determination of the landing safety buffer will result in lower runway throughputs, with larger landing intervals between landing aircraft. In extreme cases (e.g., severe convective weather conditions), the delay might be very significant and prolonged due to high congestion and weather uncertainties. Real-time traffic management systems (e.g., Integrated Arrival Departure Surface Traffic Management by NASA) consider potential conflicts by constantly adjusting the group of aircraft within TMA in terms of re-routing, re-timing, and holding \citep{sama2014optimal}.

To properly include MST as the safety-critical \textit{landing buffer time} with various operational uncertainties, we predict the ETA along with the corresponding ETA confidence interval. The prediction of ETA usually happens upon the aircraft entering the TMA, which is usually 40 minutes ahead of landing \citep{ciesielski1998real}. Early works to predict arrival time focus on using physics-based trajectory models, which are usually associated with the aircraft performance, flight plan, and the predicted atmospheric conditions provided by flight-desk systems \citep{doi:10.2514/6.2016-1852}. In \cite{krozel1999estimating}, a method is proposed to predict the arrival time in heavy weather conditions using the aircraft dynamics and weather avoidance algorithm. Estimated time of arrival time prediction is approached from a hybrid linear system in \cite{roy2006target}, then the chosen route probability is further incorporated for stochastic arrival time prediction \citep{huang2007probabilistic}. A state-dependent hybrid estimation method is used for improved prediction accuracy in \cite{wei2015estimated}. Many 4D trajectory prediction algorithms with various kinematic assumptions can also provide an estimated time of arrival \citep{porretta2008performance,xi2008simulation,doi:10.2514/6.2014-2198}.

Data-driven methods for arrival time prediction have increased rapidly in recent years, due to the rise of machine learning and well-maintained data storage facilities. Tree-based methods have been used to predict air traffic delays \citep{rebollo2012network,liu2020generalized,manna2017statistical,chakrabarty2019flight}, where the weather-related features are taken into account to enhance arrival time prediction capabilities. However, tree-based methods with quantile regression \citep{glina2012tree} have not been used for uncertainty quantification of arrival time predictions. Deep learning methods, such as recurrent neural networks (RNN), are also adopted for arrival time prediction under different circumstances \citep{ayhan2018predicting, WANG2020101840}. Moreover, the importance of feature selection in air traffic prediction is discussed in \citep{dhief2020predicting}. The experiments with Changi extended TMA conclude that when building machine learning models for air traffic prediction tasks, feature selection with the help of domain knowledge is critical. The model performance is less sensitive to the selection of machine learning algorithms itself. This also guides the discoveries on feature studies and case analysis in the later sections of this paper.

\subsection{Aircraft Landing Scheduling\label{subsec: review-als}}
The definition of the ALS problem is as follows. Assume that there are $n$ aircraft lining up for landing on a single runway. The objective of the ALS problem is to find a schedule of the respective landing time $\{t_1, t_2, \ldots, t_i\}$ for each aircraft $\{1, 2, \ldots, i\}$. In ALS, there are two constraints to be satisfied: 1) the aircraft must land within a specific time period; 2) the minimum separation time between each pair of landing aircraft should be guaranteed \citep{yua2011realytime}. The common practice for ALS used by ATC is following the First-Come-First-Served (FCFS) rule, where the scheduled landing sequence is consistent with the time for each aircraft entering the TMA. FCFS is convenient to maintain safe landing operations but can lead to severe delays during busy hours (e.g., \Cref{fig: delay-compare}). The FCFS rule has several known drawbacks, a) the lower speed leading aircraft will impact the following aircraft, even if the following aircraft has higher ground speed; b) the FCFS rule can create unnecessary long separations for aircraft with different weight classes; c) when terminal area congestion presents, the aircraft reaching TMA will hold and poses significant delays to wait for the clearance of congested aircraft; d) the delay will easily propagate to several hours in extreme scenarios. 

Thus, many researchers have proposed different approaches to optimizing the aircraft landing sequence within the scheduling range. The ALS problem can be formulated into a mixed-integer programming problem \citep{beasley2000scheduling}, where the relationship to machine scheduling problem has been exploited in the literature \citep{brentnall2006aircraft, artiouchine2008runway}. Researchers propose a variety of algorithms to address the ALS problem: 1) the ALS problem can be classified into dynamic and static scheduling approaches, depending on whether the environment is dynamically changed or not \citep{ciesielski1998real, beasley2000scheduling, beasley2004displacement, bennell2017dynamic}; 2) the scheduling algorithm itself considers various impact factors and objective functions, such as airlines' preferences \citep{soomer2008scheduling}, ground workload \citep{boysen2011scheduling}, and cellular automation \citep{yu2009real}; 3) consider the ALS problem from limited airspace or extended airspace. A detailed review of the above-mentioned three major perspectives is given below. 

\textbf{Static Scheduling v.s. Dynamic Scheduling} Static aircraft landing scheduling defines the ALS problem with a predetermined time window, such that the scheduling constraints are ensured. \cite{beasley2000scheduling} proposes a mixed-integer zero-one formulation of ALS for both single and multiple runway scenarios, to consider commonly encountered issues in practice (e.g., restricting the number of total landings in a given period). The problem is further solved with linear programming-based tree search. \cite{ding2007aircraft} proposes a static optimization algorithm for aircraft landing in a single-runway, uncontrolled airport, with performance metrics such as total holding time and total landing time. Some other researchers view dynamic programming as a feasible approach to ALS. \cite{faye2015solving} adopts \cite{beasley2000scheduling}'s formulation but with a novel dynamic constraints generation algorithm. The proposed algorithm approximates the MST into a rank two matrix, which leads to linear programming with relaxation. The dynamic ALS problem received less attention in the literature and is usually achieved with the same approach called rolling horizon \citep{bennell2017dynamic}. Rolling horizon is as simple as rolling the time window of agents for optimization. Firstly, the aircraft inside TMA within the rolling horizon (typically several minutes) are optimized. Then, the landed aircraft are removed from the rolling horizon, and the new aircraft just entered the rolling horizon are added to the algorithm. \cite{ciesielski1997anytime} solves dynamic ALS with genetic algorithms using data from Sydney airport, and shows that the genetic algorithm can perform good results in real-time with a rolling horizon of 3 minutes.

\textbf{Optimization Objectives \& Related Factors} Researchers working on the ALS problem consider various impact factors with different optimization objectives. In \cite{beasley2000scheduling}, the authors focus on reducing the deviation from the scheduled landing times. A linear programming-based tree search method was proposed for landing scheduling, building upon the pioneering work of mixed-integer programming formulation for ALS \citep{abela1993computing}. Similarly, \cite{beasley2001scheduling} extend the work to reduce deviations from scheduled landing times under time window constraints, but the MSTs are pre-defined for five different aircraft weight classes. Based on the tree search approach proposed in \cite{beasley2000scheduling}, \cite{soomer2008scheduling} considers airline preferences into the optimization framework, in which the optimal landing sequences are given by tree search and MILP is used to determine the optimal landing time. Dynamic programming-based landing sequencing method is proposed in \cite{balakrishnan2006scheduling} to maximize the runway throughput. \cite{balakrishnan2006scheduling} achieves a highly satisfactory result, but the concern on computational complexity limits the real-world applicability. Studies on alleviating computational complexity are also conducted, such as the cellular automaton optimization method \citep{yu2011real}, ant colony optimization method \citep{farah2011ant}, genetic algorithm \citep{ciesielski1998real,hu2005genetic}, and population heuristic algorithm \citep{beasley2001scheduling, pinol2006scatter}.

\textbf{TMA Scheduling Range} There have been several studies focusing on changing the range of the TMA for ALS. Some of the researchers propose to perform landing scheduling on the entire TMA, to consider the ALS problem from a systematic view. \cite{d2012aircraft,d2015real} divide the ATC controls into routing decisions, scheduling decisions, and air segments and runways. Then, a job shop formulation is used to reduce the delay caused by conflicts in TMA. More recently, researches on arrival management suggest that performing aircraft sequencing in an extended area rather than in TMA is actually an effective solution. This concept allows ATCs to monitor and control traffic into a busy terminal area from the en-route phase, enabling aircraft to adjust their speed before their top of descent. Thus, time spent in mid-air holding in the TMA can be reduced. In \cite{toratani2018study}, an algorithm is developed using the merging optimization method to simultaneously optimize trajectories, arrival sequence, and allocation of aircraft to parallel runways. A two-stage stochastic mixed-integer programming model is proposed in \cite{khassiba2020two}. Another study \cite{vanwelsenaere2018effect} assessed the effect of flights departing on extended arrival management, in terms of flight crew and air traffic control task load, sequence stability, and delay. Two-stage stochastic programming is presented in \cite{khassiba2019extended} to address the arrival sequencing and scheduling problem under uncertainty. 

\cite{ikli2021aircraft} provides a comprehensive review of optimization methods for the aircraft runway scheduling problem, covering exact methods, metaheuristics, and new approaches such as reinforcement learning. The manuscript identifies analogies with classic problems like traveling salesman and vehicle routing that provide insights. Notably, this paper constructs new challenging test instances from real air traffic data to serve as a benchmark for ALS studies. It provides a thorough overview of optimization techniques for the ALS problem and sets the stage for future research by identifying the limitations of current approaches and proposing new benchmark instances. The major limitation is the lack of uncertainty handling for real-world settings.

Several existing gaps can be identified from the above review. For example, the existing landing scheduling methods assume the actual arrival times to deviate randomly from target times (calculated using the en-route speeds) to infer MST. Also, the scheduling algorithm assumes a pre-defined MST based on the aircraft weight classes. In practice, there is tremendous uncertainty associated with the arrival time prediction, which violates the assumptions of deterministic separation. Several factors, such as aircraft type, weather conditions, and airspace density information can be explicitly acquired in the aviation database and should be used to reduce the uncertainties of arrival time prediction. In addition, the assumption of static and fixed arrival time distributions is not valid and may cause ineffective landing scheduling and/or unsafe separation between aircraft (examples shown later using realistic data). The exact arrival time prediction with accurate uncertainty quantification for each landing aircraft should be determined, which further optimizes landing schedules for all of the landing aircraft with an ensured confidence level. Thus, the main focus of this paper is to develop a real-world data-enhanced landing scheduling algorithm to achieve optimal landing scheduling with uncertainties.

\section{Methodologies\label{sec: methodologies}}
This section demonstrates the methodologies for ML-enhanced ALS. We first illustrate the tree-based machine learning selected -- Gradient Boosting Machine (GBM) with quantile regression in \Cref{subsec: review-gbm}. Then, we provide the necessary background to the Traveling Salesman Problem and introduce the formulation of time-constrained TSP formulation to solve the ALS problem in \Cref{subsec: review: tsptw}. Following this, we describe our proposed approach to integrating machine learning prediction of arrival time into time-constrained TSP formulation in \Cref{subsec: incorporating}. 

\subsection{Gradient Boosting Machine\label{subsec: review-gbm}} 
Although the literature has concluded that selecting the correct feature set is more advantageous than pursuing the most advanced machine learning algorithms \citep{dhief2020predicting}, we choose tree-based machine learning algorithms due to their proven outstanding performances on structured data \citep{grinsztajn2022tree, qin2021neural}. Furthermore, across various tree-based machine learning algorithms, we select boosting over simple trees or bagging. The benefits of boosting are threefold, (a) Boosting methods add new base learners to the ensembles at each iteration, and each base learner has trained w.r.t. the residual from the current ensembles. As a result, this iterative process helps to reduce bias and increase model accuracy. (b) Boosting provides feature importance as the indicator of critical features, which is valuable for feature selection and understanding the input-output relations. (c) Boosting can capture complex patterns by capturing complex decision boundaries over simple trees and bagging. Gradient Boosting Machine (GBM) is a commonly used model \citep{friedman2000additive,friedman2001greedy,friedman2002stochastic}. GBM connects boosting and optimization \citep{friedman2001greedy, friedman2002stochastic} to perform gradient descent on both the loss functions and the base learners. The exceptional performance of GBM on ETA prediction is also concluded by a recent study \cite{wang2020automated}.

Considering a supervised learning problem with structured data $\mathcal{D}=\{(x_i,y_i)|i=0,...,n\}$, where $x_i \in \mathbb{R}^M$ is also called the feature vector of the $i$-th sample with $M$ different features, and $y_i$ is the continuous response as the label of $x_i$ in a regression problem. In GBM, we have a set of base learners $\mathcal{B}=\{b_{\gamma_m}(x) \in \mathbb{R}^M\}$ parameterized by $\gamma_m$. In GBM, the predictions are the linear combination $\mathsf{lin}\{B\}$ of the predictions from each of the base learner $b_{\gamma_m}(x) \in \mathcal{B}$, where each of the base learners is additively learned with pseudo-residuals ($\tau_m$). The corresponding predicted label to feature vector $x_i$ is $f(x_i) \in \mathsf{lin}\{B\}$, in the form of,

\begin{equation}
    f(x_i) = \sum^M_{m=0} \alpha_m b_{\gamma_m}(x_i)
\end{equation}

\noindent where $\alpha_m$ is the coefficient for each base learner $b_{\gamma_m}(x) \in \mathcal{B}$. Examples of base learners include linear models, support vector machines, classification, and regression trees \citep{friedman2001greedy,lu2020accelerating}. Additionally, for the most popular tree-based learners set, the GBM turns into Gradient Boosted Decision Trees (GBDTs). GBM aims to obtain the best function set $\hat{f}(x_i) \in \mathsf{lin}\{B\}$ to minimize the given data-fidelity evaluation function (e.g., least squared residual loss for simple regressions).

\begin{equation}\label{eq: gbm_loss}
    \hat{f}(x_i) = \mathsf{argmin}_{f(x_i) \in \mathsf{lin}\{B\}}  \sum_{i=0}^n \xi_i(y_i, f(x_i))   
\end{equation}

\noindent where $\xi_i(y_i, f(x_i))$ is the data-fidelity evaluated at the $i$-th feature vector. 

Using the defined notations above, the GBM minimizes the loss function by calculating the steepest descent to the objective function defined in \Cref{eq: gbm_loss}, where the steepest gradient is determined with line-search on the best base learner parameter set $\hat{\gamma}_m$.



The following research explores the possibility of improving the boosting method performance from many perspectives \citep{friedman2002stochastic}.
\begin{itemize}
    \item Introduce a learning rate $\lambda$ to the updating equation of $f(x)$: $f^{m+1}(x) = f^m(x)+\lambda \rho_m b_{\gamma_m}(x)$. Multiplying $\lambda$ provides the damping of controlling the rate of descent on the error surface.
    \item Sampling without replacement from the dataset before the gradient calculation step gives stochasticity to GBM, and greatly improved the performance of the algorithm.
    \item Using ANOVA decomposition can restrict the depth of the trees, which further controls the order of approximations of GBM: $f(x) = \sum_i f_i(x_i) + \sum_{ij}f_{ij}(x_i, x_j) + \sum_{ijk}f_{ijk}(x_i, x_j, x_k) + \cdots$ 
\end{itemize}

Specifically, GBMs can be turned into probabilistic predictors when applying quantile distributions to the response variable. The gradient calculation of $\tau_m$ changes to the quantile pseudo-residual $\tau_m = \beta \xi(y_i \geq f(x_i)) - (1-\beta) \xi(y_i \leq f(x_i))$, where $\beta = \frac{\sum_{i=0}^n \omega_i \xi_i(y_i \leq q)}{\sum_{i=0}^n \omega_i}$ and $q$ denotes the weighted quantile. With the GBM prediction label of the defined quantile, given a test sample $\hat{x_i}$, we can obtain the confidence interval $\sigma$ along with the prediction $\hat{y_i}$.

\subsection{Traveling Salesman Problem with Time Windows (TSP-TW)\label{subsec: review: tsptw}}
The ALS problem involves landing sequencing and landing scheduling, which is a discrete optimization problem in nature. Combinatorial optimization tackles discrete optimization problems from the intersection of combinatorics and theoretical computer science. Combinatorial optimization is widely utilized to solve tasks like resource allocation and scheduling for transportation and supply chains. TSP-TW is a classical combinatorial optimization problem \citep{alharbi2021solving}. The original definition of TSP-TW aims at finding the optimal tour that minimizes the length of the tour, and visits each node once within the specified time window $[l_i, u_i]$, where $l_i$ and $u_i$ are the lower and upper bound for visiting time of node $i$. The bounded time windows set time constraints to the agent traveling within the node graph, and mark the significant difference to classical TSP problems. TSP-TW has been applied to bus scheduling and delivery systems \citep{alharbi2021solving}. In this work, we propose to use TSP-TW for ALS, and incorporate the machine learning predicted aircraft ETAs into the constraints of TSP-TW.

Define an undirected graph $G = (V, A)$ with a finite set of nodes, $V=\{0,1,\dots,n\}$, and a finite set of edges, $A = \{(i,j)|i\neq j, i,j \in V\}$. TSP-TW determines the time $t_i$ that the agent visits node $i \in \{0,1,\dots,n\}$. Meanwhile, an additional variable, $t_{n+1}$, is introduced to represent the completion time of the tour, as the agent has to return to node 0 at the end of the tour. A distance matrix, $t_{ij}$, records the shortest distance between each node pair, which can be further treated as the scalar transformation of time distance between node pairs \citep{baker1983exact}. Mathematically, the classical formulation of TSP-TW is shown in \Cref{eq: tsp-tw,tsp_st1,tsp_st2,tsp_st3,tsp_st4,tsp_st5}. 

\begin{equation}\label{eq: tsp-tw}
    \min \quad t_{n+1} 
\end{equation}
subject to
\begin{align}
    t_i - t_0 &\geq t_{0i}       &&i=1,2,\dots,n \label{tsp_st1}\\
    |t_i - t_j| &\geq t_{ij}  &&i=2,3,\dots,n; 1 \leq j < i \label{tsp_st2}\\
    t_{n+1} - t_i &\geq t_{i0}&&i=1,2,\dots,n \label{tsp_st3}\\
    t_i &\geq 0               &&i=0,1,\dots,n+1 \label{tsp_st4}\\
    l_i &\leq t_i \leq u_i    &&i=1,2,\dots,n \label{tsp_st5}
\end{align}

To solve TSP-TW, there are several methods spanning from mathematical programming approaches to heuristic approaches. Mixed-Integer Linear Programming (MILP) techniques are commonly used approaches to solve TSP-TW. Despite the difference between problem setups and applications, researchers propose various methods to solve TSP-TW with MILP for up to 200 clients \citep{christofides1981state,baker1983exact,dumas1995optimal}. Additionally, constraint programming methods are proposed to develop both exact \citep{pesant1998exact} and heuristic \citep{pesant1999flexibility} solvers for TSP-TW. While in this work, we incorporate the MST into the constraints (\Cref{st1,st2,st3,st4,st5,st6,st7,st8}) of the TSP-TW model, keeping the original objective function in \Cref{eq: tsp-tw}.

\begin{align}
    t_i &\geq t_{0i} \cdot y_{0i}&& i=1,2, \dots,n  \label{st1} \\
    t_i-t_j + (u_i-l_j+t_{ij})\cdot y_{ij} &\leq u_i - l_j &&\forall i,j=1,2,\dots,n : i\neq j  \label{st2}  \\
    \sum_{i=0}^{n}y_{ij} &= 1&& j=1,2, \dots,n   \label{st3} \\
    \sum_{j=0}^{n}y_{ij} &= 1&& i=1,2, \dots,n  \label{st4}  \\
    t_i +t_{i0} &\leq t_{n+1} && i=1,2, \dots,n  \label{st5} \\
    l_i\leq t_i &\leq u_i && i=1,2,\dots,n  \label{st6}  \\
    y_{ij} &\in \left\lbrace 0,1\right\rbrace && \forall (i,j)\in \left \lbrace(i, j): i,j\in  {0, 1, ..., n} \right\rbrace  \label{st7}  \\
    t_i &\geq 0 && \forall i=0,1,...,n+1  \label{st8} 
\end{align}

The objective is to minimize the total landing time for all landing aircraft. $u_i$ and $l_i$ denote the earliest and latest time for aircraft $i$ to land, respectively. $u_i - l_i$ indicates the maximum allowed flight time of aircraft $i$, which can reflect the aircraft conditions (e.g., fuel, pilot fatigue level, etc). $y_{ij}$ is the adjacency matrix, defined as,

\begin{equation}
    y_{ij}=
\bigg \{
\begin{array}{ll}
1,\quad \text{if aircraft $j$ lands right after aircraft $i$,}\\
0,\quad \text{otherwise.} \\
\end{array}
\end{equation}

\Cref{st2} describes the constraint on the separation requirement between two consecutive intermediate aircraft. $t_{ij}$ denotes the MST from aircraft $i$ to aircraft $j$. \Cref{subsec: incorporating} will discuss the method used to incorporate GBM predicted MST into $t_{ij}$. As MST depends on the wake turbulence generated by the leading aircraft, the formulation is an asymmetric TSP-TW problem, indicating $t_{ij} \ne t_{ji}$. The time window for the agent to visit a node corresponds to the specified time range for the aircraft to start to land, considering the fuel consumption and aircraft dynamics. \Cref{st1,st5} guarantees that the smallest and largest $t_i$ values. \Cref{st3,st4,st7} ensure that each aircraft will land exactly once. \Cref{st6} introduces the pre-determined time schedule of each aircraft. In practice, we solve the model in \Cref{eq: tsp-tw} with GLPK solver \citep{Mak01} and Python Optimization Modeling Objects (Pyomo) package \citep{hart2017pyomo}.

\subsection{Incorporating Uncertainties of MST Constraints to TSP-TW\label{subsec: incorporating}}
As we reviewed earlier, there are tremendous uncertainties associated with the estimated arrival time and minimum separation time. Thus, the proposed study will include uncertainties in the landing scheduling problem to ensure confidence. For each successive landing aircraft pair $(i,j)$, GBM with weighted quantile gives the predicted landing time distributions since entering the extended TMA for variable $t_i$ and $t_j$ from real-world data.

For each successive landing aircraft pair $(i,j)$, GBM with weighted quantile gives the predicted landing time distributions for variable $t_i$ and $t_j$ from real-world data. It is assumed that arrival time for landing aircraft follows i.i.d. Gaussian distributions. The MST is defined as the difference between the two arrival time for the two aircraft $(i,j)$. Thus, the MST can be expressed 
\begin{equation}\label{eq: t-t-re}
    t_{ij} \sim \mathcal{N}\left(\mathcal{T}_{ij},\sigma_{ij}\right)
\end{equation}

\noindent where $\mathcal{T}_{ij}$ is the referenced MST between aircraft $i$ and $j$ by the related authorities \citep{erzberger2016algorithms}. $\sigma_{ij} = \sqrt{\sigma_i^2+\sigma_j^2}$ represents the uncertainty of MST from the quantified uncertainties (standard deviation) from the arrival time of the two aircraft $(i,j)$. The major reference values of $\mathcal{T}_{ij}$ are listed in \Cref{table: mst-table}.

\begin{table}[H]
\caption{$\mathcal{T}_{ij}$: Minimum required time-space (s) used by arrival manager \citep{erzberger2016algorithms}.}
\label{table: mst-table}
\centering
\begin{tabular}{clllll}
\hline
\multicolumn{1}{l}{}                       & \multicolumn{5}{l}{Trailing Aircraft}                                    \\ \cline{2-6} 
\multirow{5}{*}{Leading Aircraft} &       & Heavy & B757 & Large & Small \\
                                           & Heavy & 82             & 118           & 118            & 150            \\
                                           & B757  & 60             & 64            & 64             & 94             \\
                                           & Large & 60             & 64            & 64             & 94             \\
                                           & Small & 60             & 64            & 64             & 94             \\ \hline
\end{tabular}
\end{table}

Given a fixed spacing conflict probability $P_c$, the MST between landing aircraft $i$ and $j$, $\Check{\Check{t_{ij}}}$ can be calculated,
\begin{equation}\label{eq: inverse}
    \Check{\Check{t_{ij}}}=\Phi^{-1}_{t_{ij}}(P_c)
\end{equation}

\noindent where $\Check{\Check{t_{ij}}}$ forms the separation constraints in \Cref{eq: tsp-tw}. By \Cref{eq: inverse}, the minimum allowable separation time between two successive landing aircraft pair $(i,j)$ is obtained as $\Check{\Check{t_{ij}}}$. It is worth pointing out that $\Check{\Check{t_{ij}}}$ is different from $\Check{\Check{t_{ji}}}$, since the MSTs are significantly impacted by the leading aircraft. Additionally, the predicted mean values of estimated landing times $\mu_i$ and $\mu_j$ are included in the upper and lower bound ($u_i$ and $l_i$) of \Cref{st2} and \Cref{st6}. 

In this work, we aim to predict the arrival time from 200 miles of the TMA, and the aircraft can adjust the speed in the en-route phase to reach the scheduled arrival time. The fuel consumption can be limited to a low level if the scheduled arrival time is constrained to a time window around the optimal speed. Thus, fuel consumption is also considered in the constraints. We incorporate the fuel consumption constraints into the calculation of upper bound $u_i$ and lower bound $l_i$ of the time window constraints, adjusted based on the distribution of $t_{ij}$. 

\begin{figure}[H]\label{fig: als-framework}
    \centering
    \includegraphics[width=0.85\textwidth]{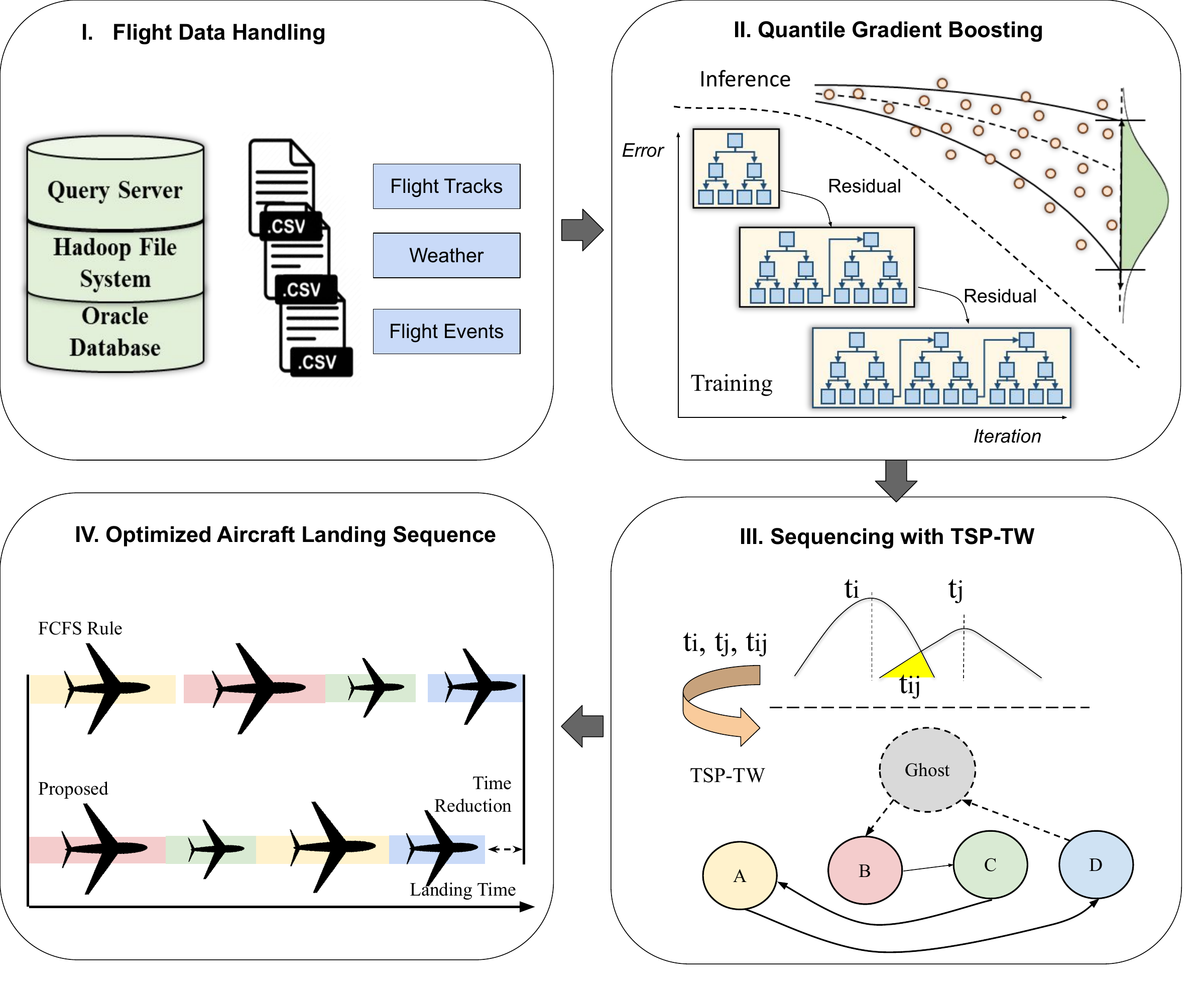}
    \caption{Overview of the proposed machine learning-enhanced optimization model for aircraft landing scheduling. }
\end{figure}

The complete ALS procedure is shown in \Cref{fig: als-framework}. The aviation source data are obtained from the Sherlock Data Warehouse and processed into well-organized feature representations. The Boosting model takes the processed feature and fits into base learners sequentially, where the residuals are concatenated for the best set of base learners. The boosting model predicts the distribution of the landing time for each landing aircraft $t_i$, $t_j$. $t_{ij}$ is further derived as the constraint of TSP-TW for optimal landing sequencing. GBM is trained with real-world air traffic data, to predict the estimated aircraft landing time with associated uncertainty intervals. First, a look-ahead horizon is defined to determine the rolling time window for scheduling. For instance, the TMA ranges from 100 nautical miles to 200 nautical miles from the destination airport. The detected number of aircraft in this area is $n$. The maximum number of aircraft to be scheduled is constrained to $n_{max}$ to limit the computation complexity for real-time implementation. If $n \ge n_{max}$, the first $n_{max}$ aircraft is selected and performs the landing scheduling, otherwise select all the aircraft in the current horizon. The trained model is used to predict the arrival time of all the affected aircraft. Then the scheduled arrival time is calculated using the algorithm described above.

\section{Empirical Data Analysis\label{sec: investigations}}
In this section, we first investigate several flight delay scenarios via real-world aviation flight recordings. Through the investigations, we discover that the holding pattern is one of the major impact factors leading to flight delays. We propose to explicitly include safety-related flight event counts as the features for GBM to demonstrate effectiveness. A short description of the flight track and flight event data is given.

\subsection{Investigation on Flight Delays\label{subsec: investigations}}
\begin{figure}[H]
    \centering
    \begin{subfigure}[t]{0.485\textwidth}
        \centering
        \includegraphics[width=\textwidth]{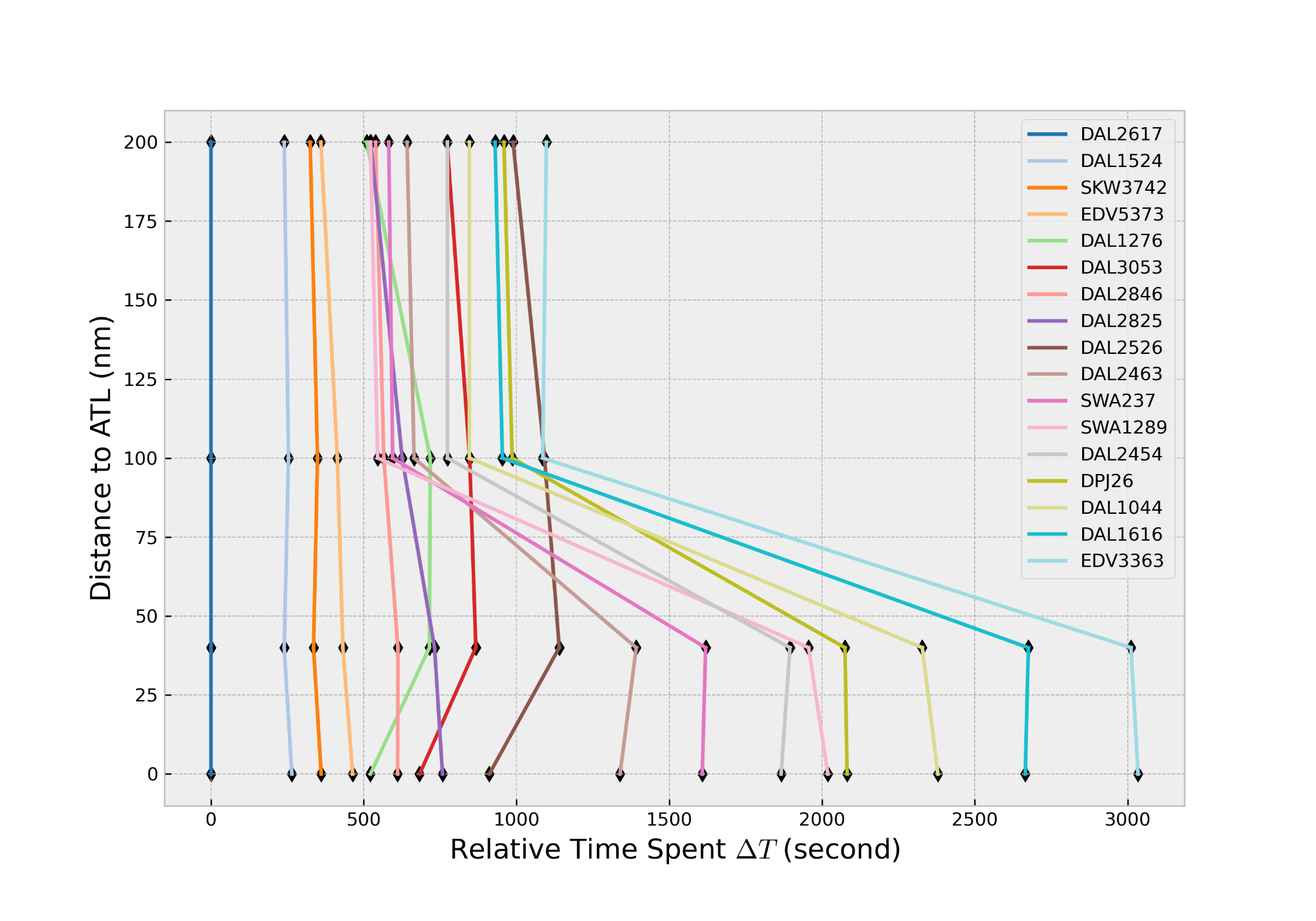}
        \caption{Aug 1st, 2019@13:20 - 13:45}
        \label{fig: 20190801}
    \end{subfigure}
    ~
    \begin{subfigure}[t]{0.485\textwidth}
        \centering
        \includegraphics[width=\textwidth]{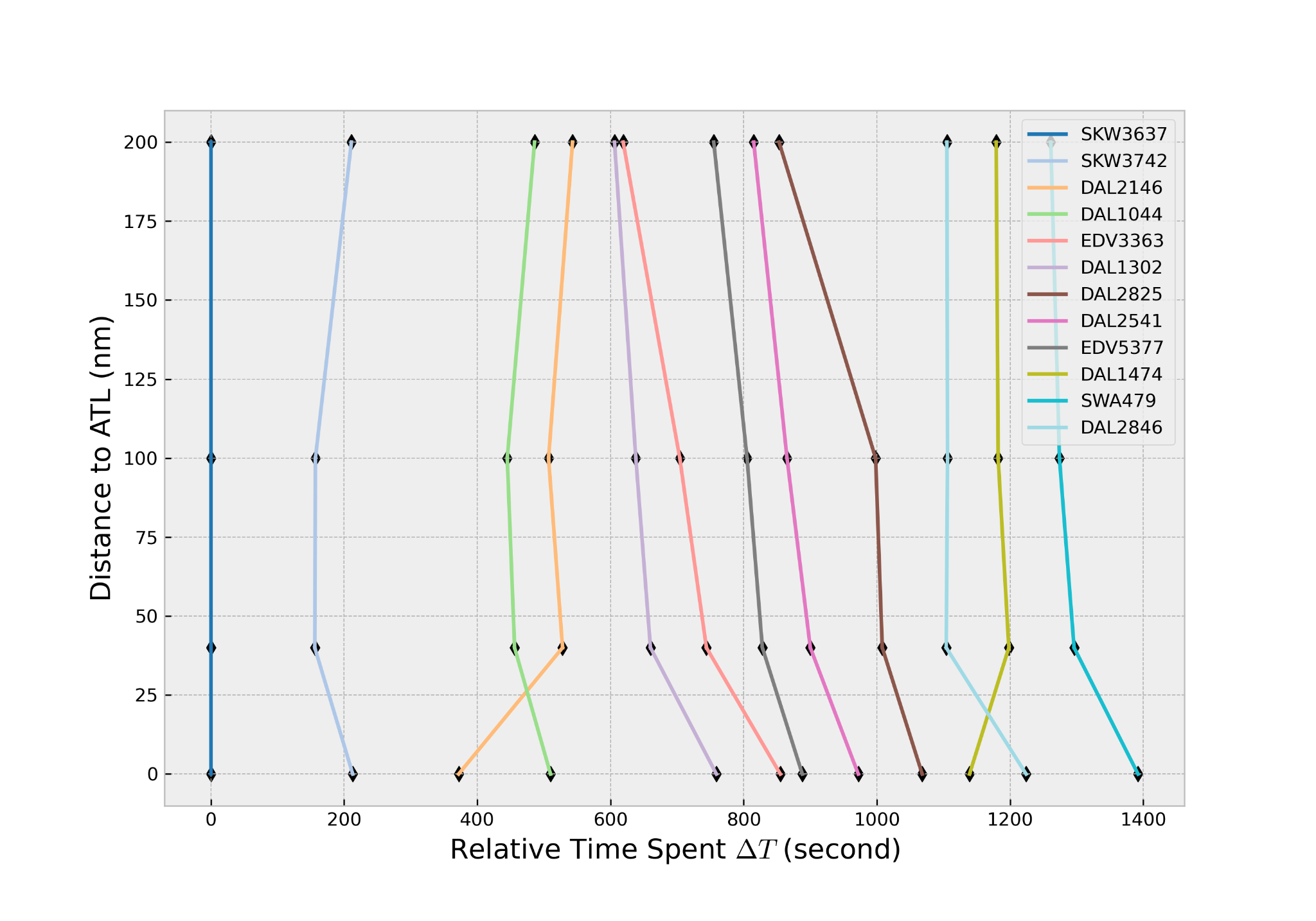}
        \caption{Aug 8th, 2019@13:20 - 13:45}
        \label{fig: 20190808}
    \end{subfigure}
    \caption{Flight landing progress recording during busy hours. A comparison between two consecutive Mondays. Both days present normal weather conditions.}
    \label{fig: delay-compare}
\end{figure}

\Cref{fig: delay-compare} shows the relevant time spent from three different distance intervals ($[200, 100], [100, 40], [40, 0]$ nautical miles) away from the terminal. We obtain and visualize the flight track data from Sherlock Data Warehouse (SDW) \citep{arneson2019sherlock}. SDW is a distributed big data platform for data visualization to support air traffic management research. Sherlock includes a database, a web-based user interface, a few data visualization tools, and other services. The flight data is stored in the Integrated Flight Format (IFF). It includes all raw data plus the derived fields such as flight summary, track points, and flight plan. The flight summary is a general description of the flight which contains flight time, flight call sign, aircraft type, origin, and destination information. The flight track points are the record of real flight operations. It includes the ground-measured aircraft position in both the spatial and temporal aspects. The format of flight track data in SDW and conversion between WGS84 coordinates to absolute distance has been discussed in previous work \citep{pang2022bayesian}. 

In \Cref{fig: delay-compare}(a), we visualize the landing aircraft between the time window $13:20$ and $13:45$ on Monday, Aug 1st, 2019, during which severe landing flight delays happened. For reference, we also visualize the normal flight landing process of the next consecutive Monday in \Cref{fig: delay-compare}(b). During the given time window, 17 flights entering the 100 nautical mile range from the terminal are captured by the surveillance radar. \Cref{fig: delay-compare}(a) shows flight delay starts at the $5$-th flight. 

\begin{figure}[H]
    \centering
    \begin{subfigure}[t]{0.25\textwidth}
        \centering
        \includegraphics[height=6cm]{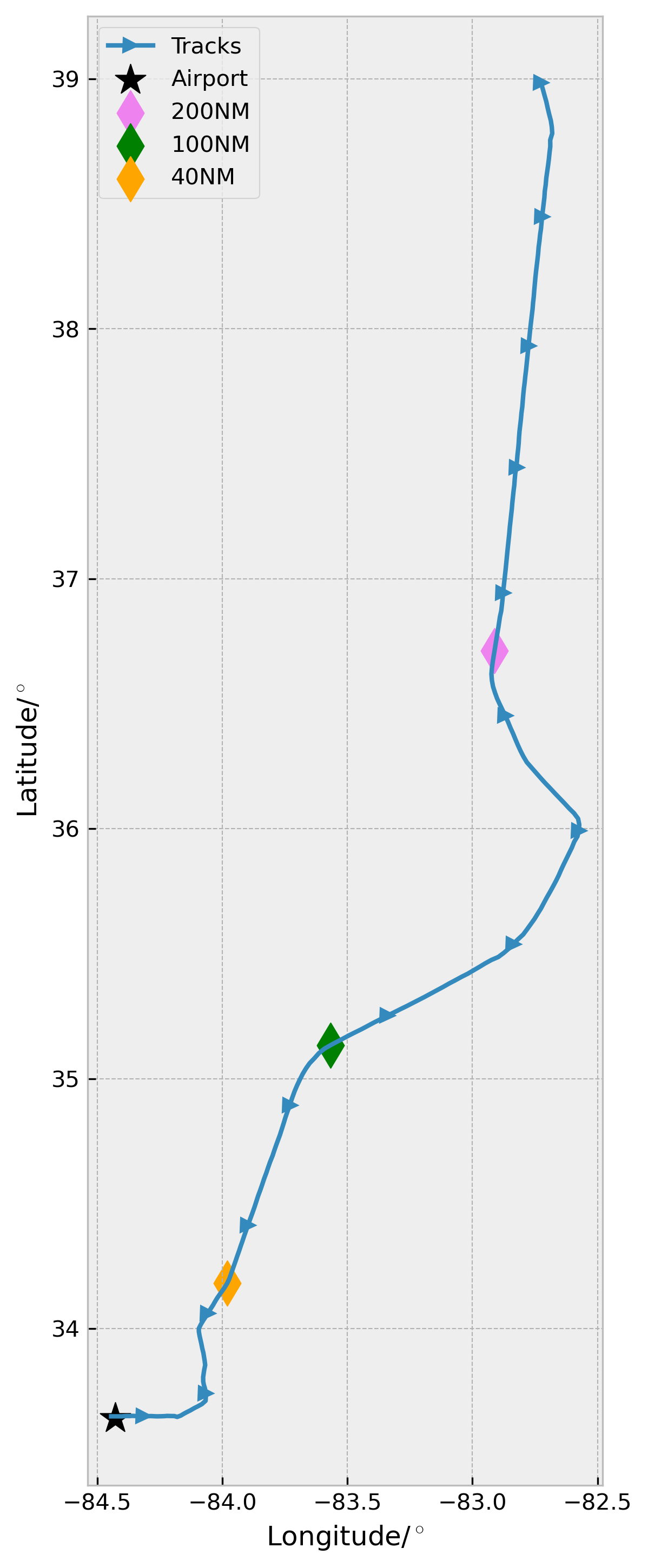}
        \caption{DAL1276}
        \label{fig: DAL1276}
    \end{subfigure}
    ~
    \begin{subfigure}[t]{0.4\textwidth}
        \centering
        \includegraphics[height=6cm]{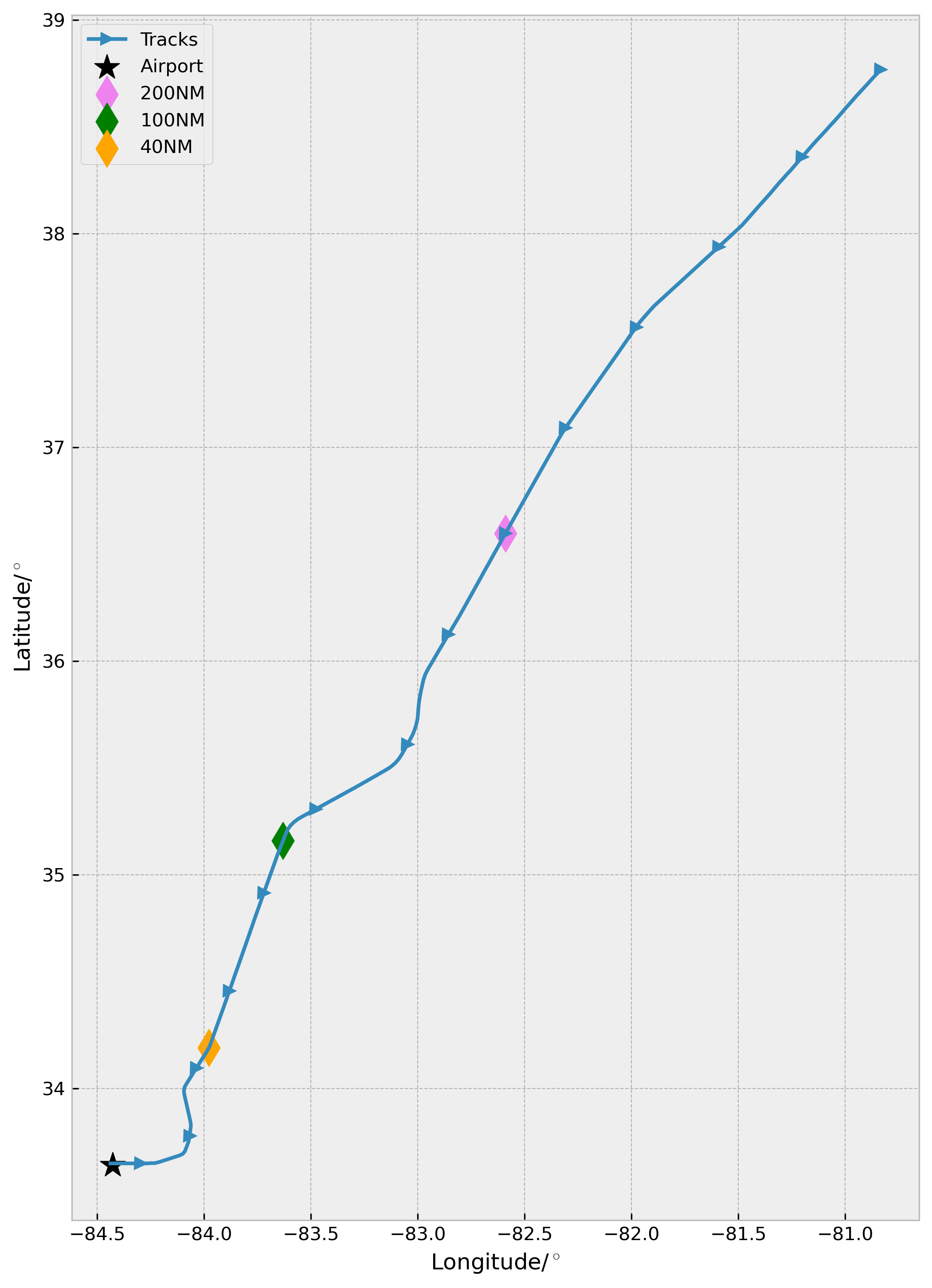}
        \caption{DAL3053}
        \label{fig: DAL3053}
    \end{subfigure}
    ~
    \begin{subfigure}[t]{0.28\textwidth}
        \centering
        \includegraphics[height=6cm]{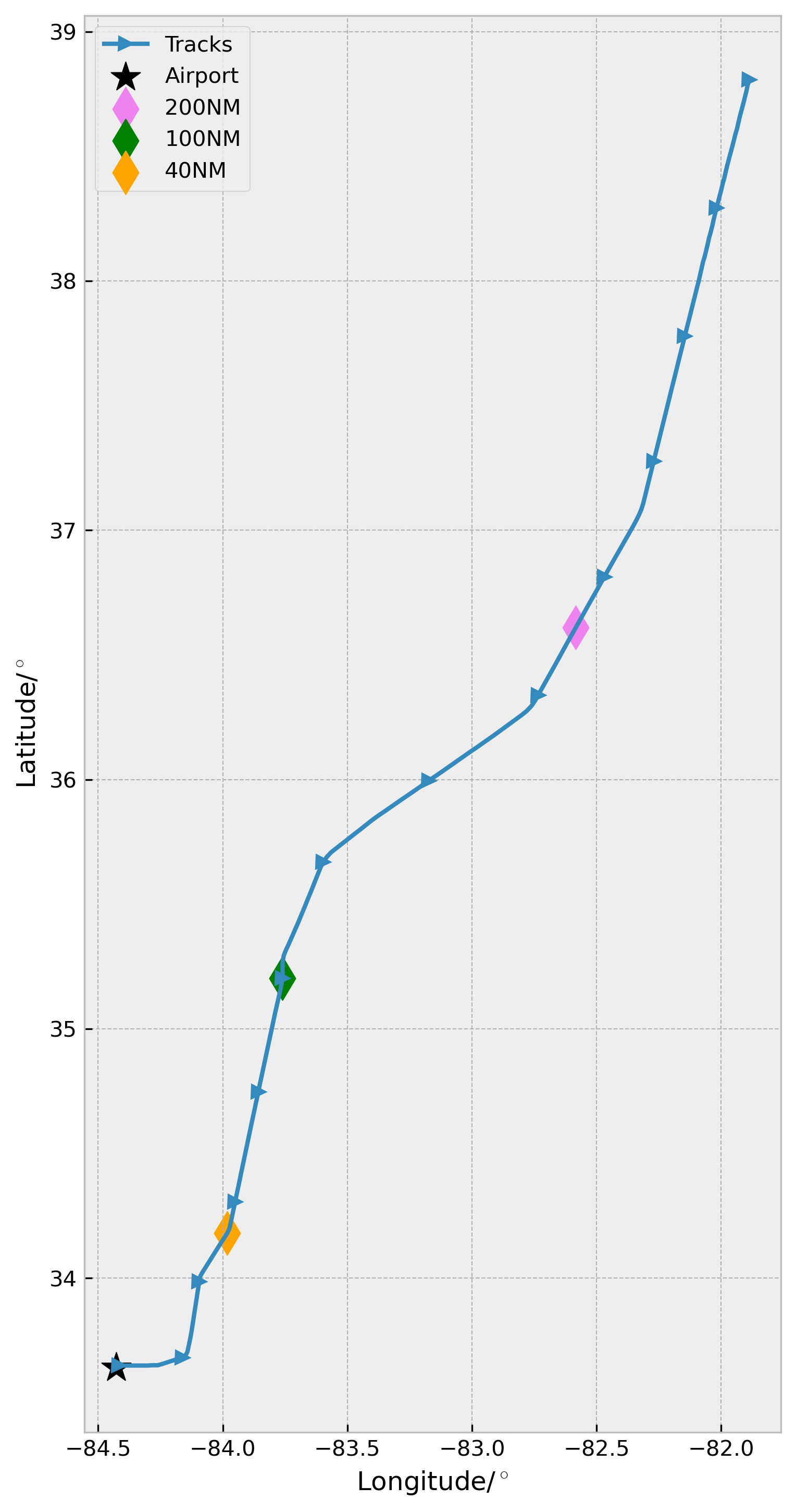}
        \caption{DAL2526}
        \label{fig: DAL2526}
    \end{subfigure}
    \caption{Landing aircraft coming from the northeast for west landing.}
    \label{fig: delay-short}
\end{figure}

\begin{figure}[H]
    \centering
    \begin{subfigure}[t]{0.9\textwidth}
        \centering
        \includegraphics[width=\textwidth,height=4cm]{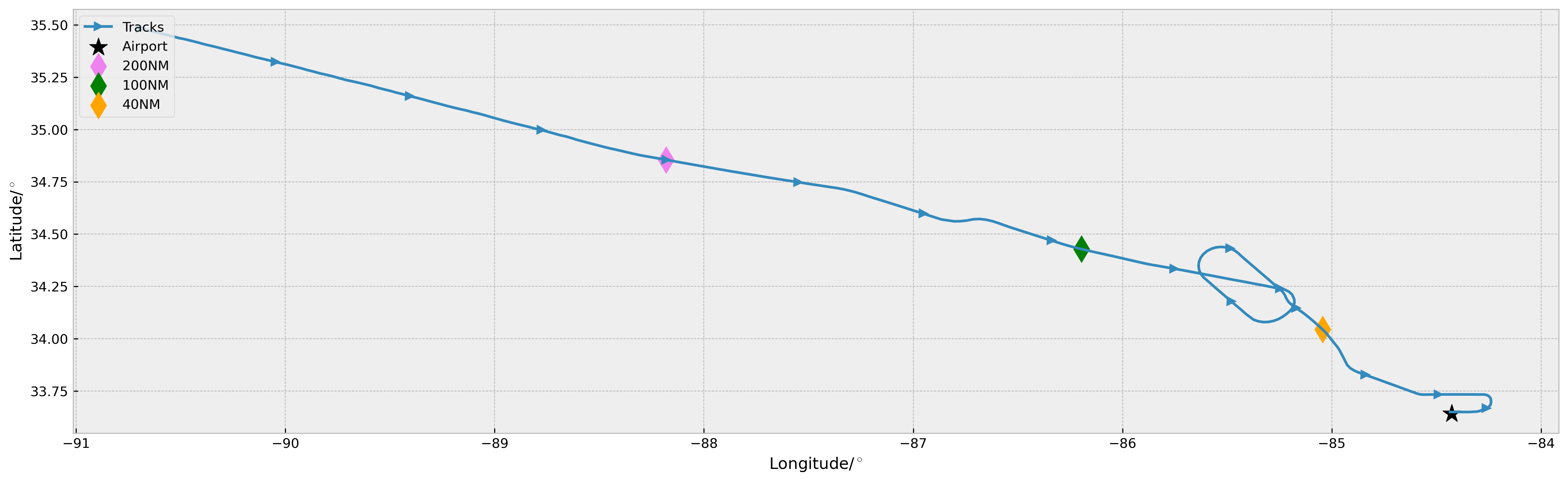}
        \caption{DAL2463}
        \label{fig: DAL2463}
    \end{subfigure}
    ~
    \begin{subfigure}[t]{0.9\textwidth}
        \centering
        \includegraphics[width=\textwidth,height=4cm]{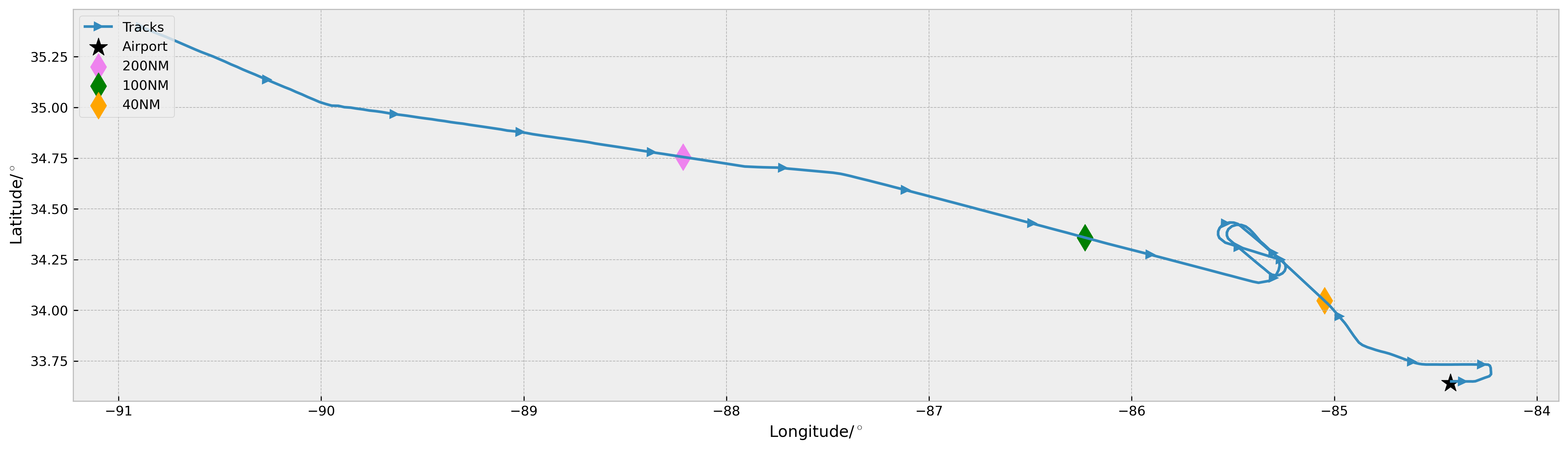}
        \caption{SWA237}
        \label{fig: SWA237}
    \end{subfigure}
    ~
    \begin{subfigure}[t]{0.9\textwidth}
        \centering
        \includegraphics[width=\textwidth,height=4cm]{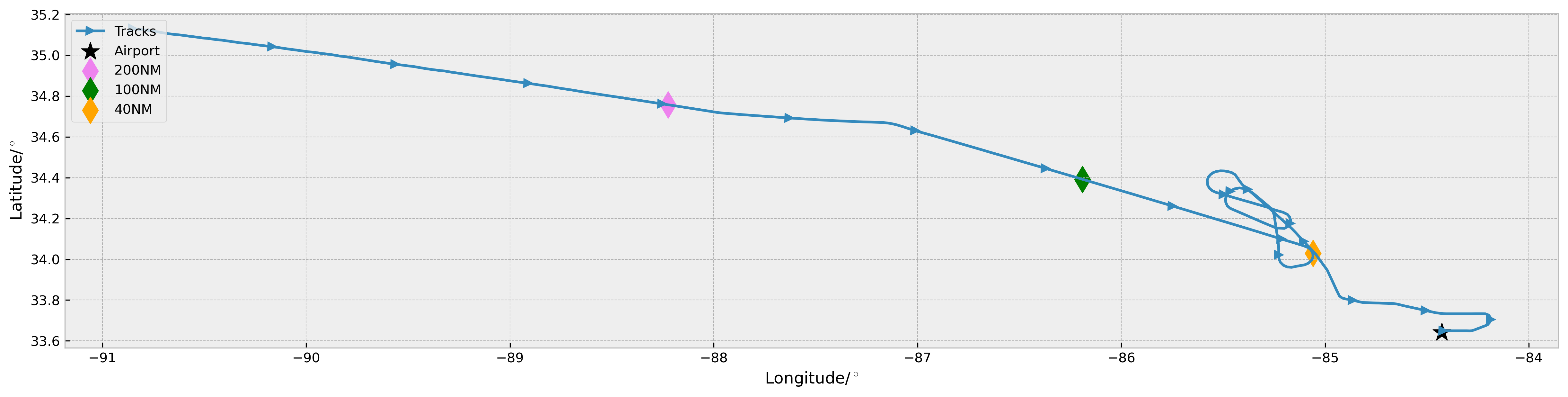}
        \caption{SWA1289}
        \label{fig: SWA1289}
    \end{subfigure}
    ~
    \begin{subfigure}[t]{0.9\textwidth}
    \centering
    \includegraphics[width=\textwidth,height=4cm]{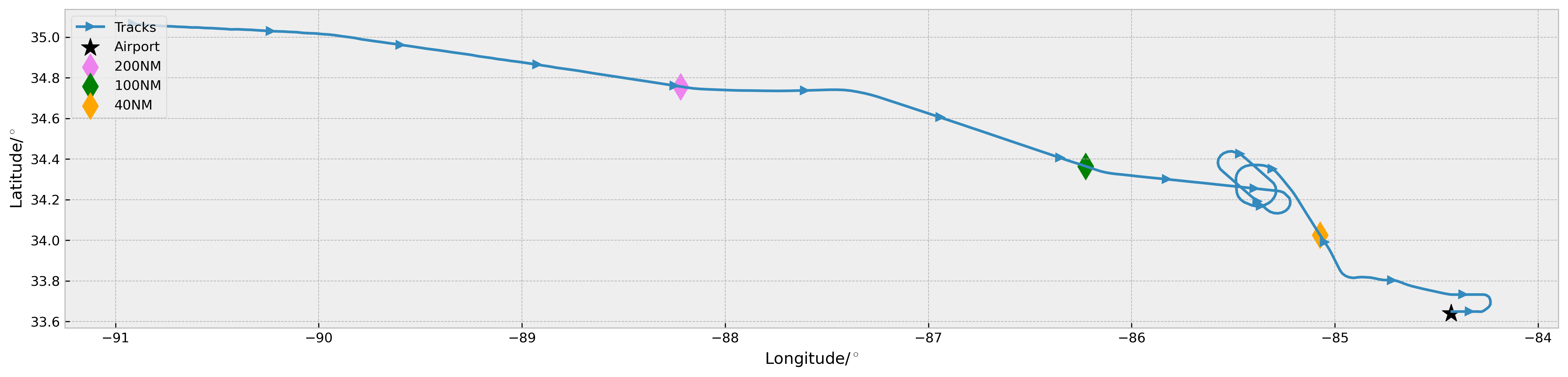}
    \caption{DAL2454}
    \label{fig: DAL2454}
    \end{subfigure}
    \caption{Landing aircraft coming from the northwest for west landing.}
    \label{fig: delay-long}
\end{figure}

We draw the complete tracks for 3 landing flights (DAL1276, DAL3053, DAL2526) coming from the northeast towards an east landing, and 7 landing flights coming from the northwest towards an east landing. In \Cref{fig: delay-short} and \Cref{fig: delay-long}, the diamond marks represent the location of reaching $200, 100, 40$ NM from the airport center. We gain valuable insights from \Cref{fig: delay-short} and \Cref{fig: delay-long}, a) holding command is the common practice during ALS, and the flight receives a holding command loop around in the airspace; b) holding pattern contributes to the long arrival time within the TMA, and can last for various period; c) holding patter usually happens when the aircraft is in TMA range $[100, 40]$NM, where FCFS rule takes effect for air traffic control.  

\begin{figure}[H]
    \centering
    \includegraphics[width=0.85\textwidth]{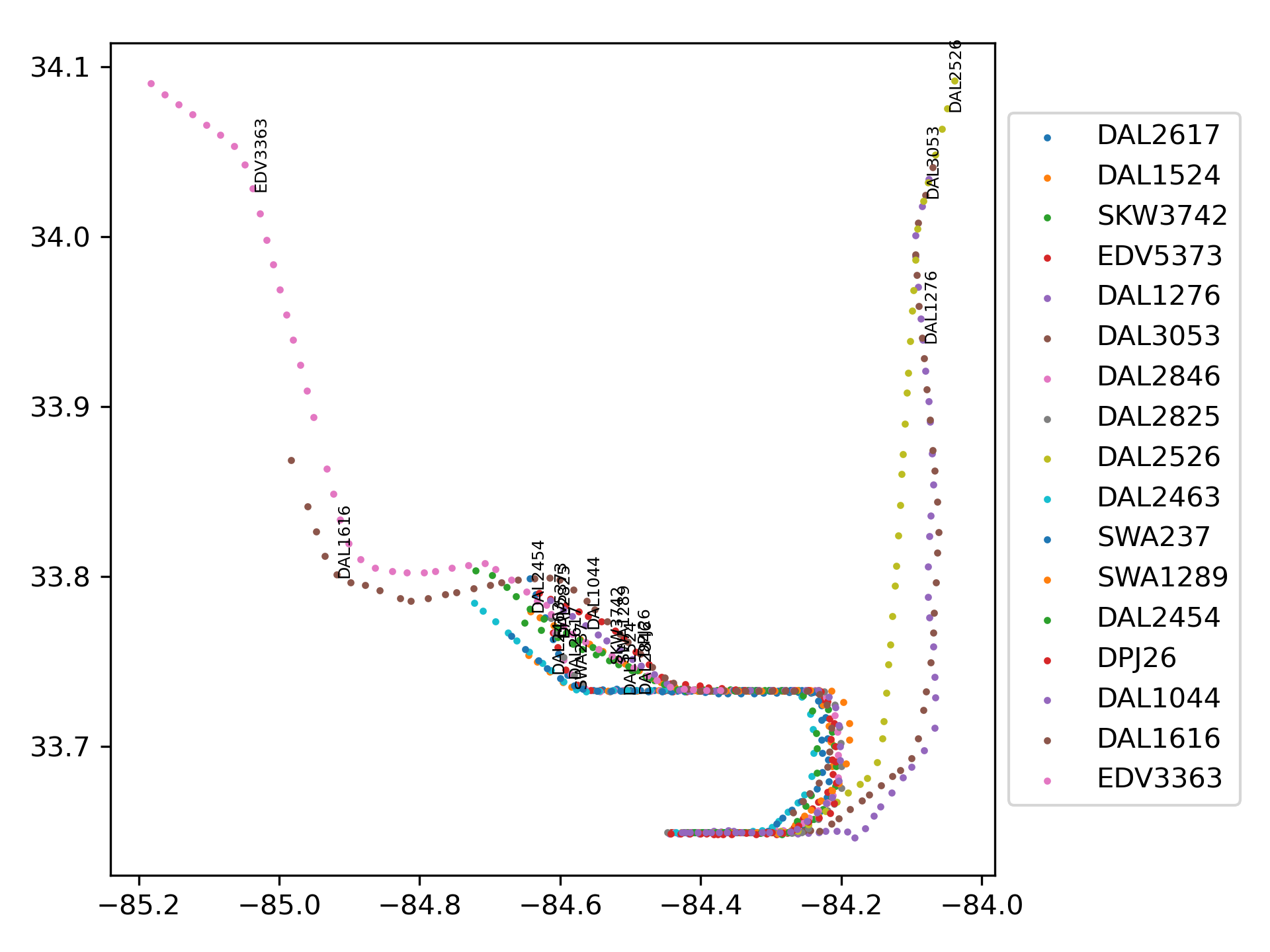}
    \caption{Flight tracks for landing aircraft in \Cref{fig: delay-compare}(a). }
    \label{fig: visualization_traj}
\end{figure}

It's obvious that holding in the congested near terminal airspace poses a safety concern to air traffic operations, which motivates our proposed method to perform from an extended TMA. In this work, we propose to do ALS from an extended TMA, such that an early landing scheduling can be issued. By doing this, we can alleviate the near terminal airspace complexity, and landing aircraft can adjust the speed profile to account for the issued arrival time, over $100$NM away from the terminal.

\subsection{Aviation Data Mining}
The aviation data used in this work are obtained from the SDW, where the flight tracks and flight event recordings are of interest, while the weather data are obtained from the open dataset.

\subsubsection{Flight Track Recordings \label{subsubsec: iff-track}}
\begin{figure}[H]
    \centering
    \includegraphics[width=0.99\textwidth]{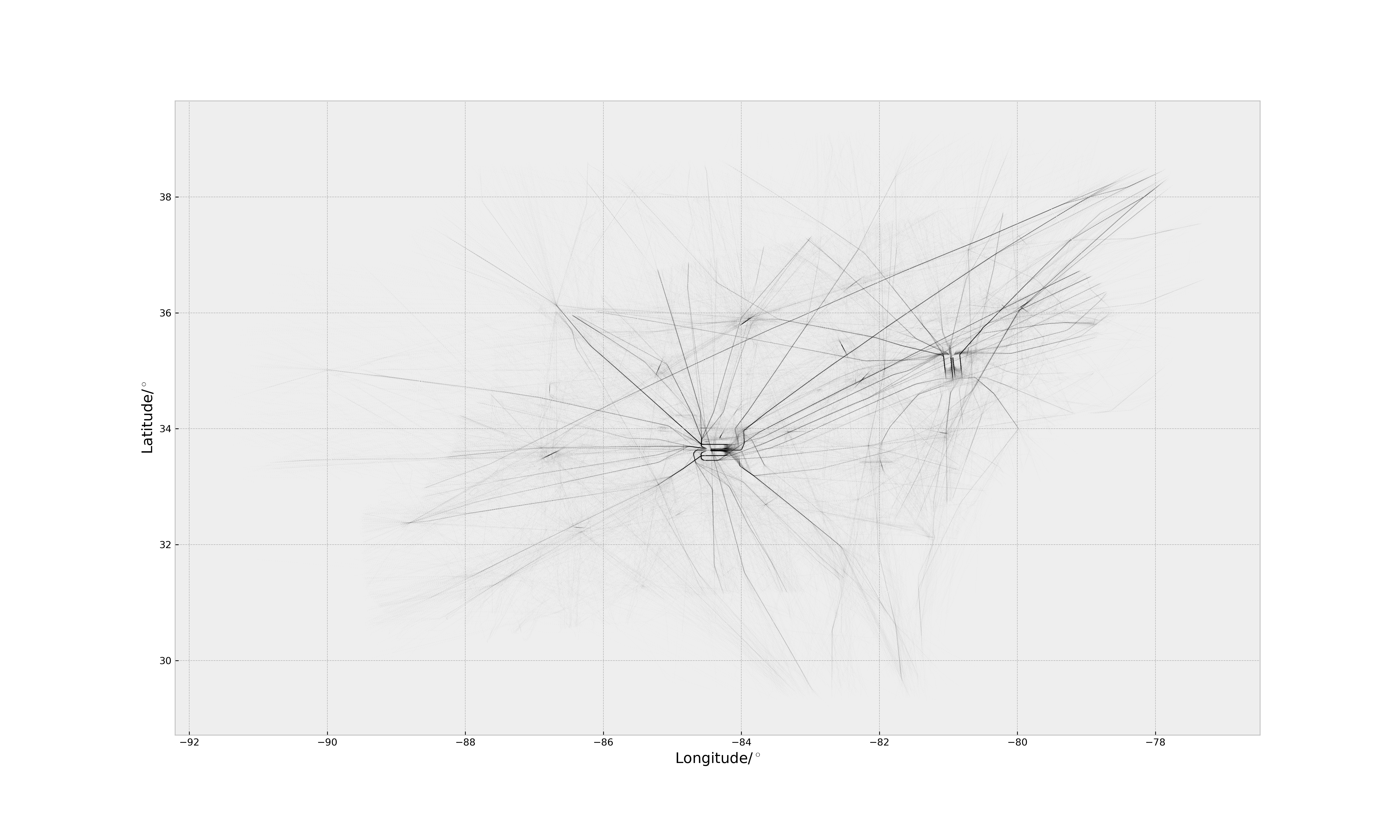}
    \caption{Visualization of IFF flight track data from the FAA Atlanta Air Route Traffic Control Center (ARTCC ZTL) on Aug 1st, 2019 obtained from Sherlock Data Warehouse \citep{arneson2019sherlock}.}
    \label{fig: plot of tracks}
\end{figure}

The flight track data takes the standard Integrated File Format (IFF) for aviation standards. The IFF flight track data contains the processed raw flight data collected from FAA facilities across the United States territories, as well as some derived features such as flight summary. We use IFF flight track data from the FAA Atlanta Air Route Traffic Control Center (ARTCC ZTL). ARTCC ZTL covers airspace across Alabama, Georgia, South Carolina, Tennessee, and North Carolina. For a better illustration of ARTCC ATL, \Cref{fig: plot of tracks} shows the flight tracks recorded in ARTCC ZTL.

\begin{table}[h]
\caption{List of Features used in Probabilistic Estimated Arrival Time Prediction Model}
\label{table: features}
\centering
\begin{adjustbox}{max width=\textwidth}

\begin{tabular}{l|l|l}
\hline
\rowcolor[HTML]{EFEFEF}
\textbf{Feature Group}              & \textbf{Feature Name}                   & \textbf{Descriptions}                                                     \\ \hline
\multirow{8}{*}{Flight Conditions}                     & acType             & Type of the aircraft                                                \\ 
              & Latitude                  & Latitude                                     \\
                          & Longitude                 & Longitude                                    \\
                          & Altitude                  & Altitude                                    \\
                          & Distance                  & Distance to the destination                                      \\ 
& Time            &  Timestamp                                          \\
                          & Hour                 & Round to the nearest full hour                  \\ 
                   & GroundSpeed              & Aircraft ground speed                        \\ \hline
\multirow{5}{*}{Airspace Complexities} & AC\_10 (30, 60)mins\_ahead  & Aircraft count 10 (30, 60) mins ahead \\
                          & AC\_10 (30, 60)mins\_behind & Aircraft count 10 (30, 60) mins behind    \\
                          & EV\_RRT\_10 (30, 60)  & Reroute event count 10 (30, 60) mins ahead \\
                          & EV\_LOOP\_10 (30, 60)  & Looping event count 10 (30, 60) mins ahead    \\
                          & EV\_GOA\_10 (30, 60) & Go around event count 10 (30, 60) mins ahead \\
\hline
\multirow{5}{*}{Weather Conditions} & windspeed &  Wind speed \\
                          & winddir &   Wind direction from the true north\\
                          & cloudcover  &  Height of the cloud cover \\
                          & visibility  &  Visibility level \\
                          & humidity  &  Humidity percentage\\
                          \hline

\end{tabular}
\end{adjustbox}
\end{table}

IFF flight track data contains the flight operational features (e.g., flight plans, flight callsign), positional features (e.g., coordinates, speed, course), and flight identifiers/codes (e.g., Beacon code, operations type). As discussed in previous sections, we are interested in the features that can represent the status of the target aircraft, as well as the nearby airspace complexity. We select and construct a proper set of features for the prediction of landing aircraft arrival times, as shown in \Cref{table: features}. These features show an impact on the prediction performances. The aircraft type is obtained from the Sherlock data. We use latitude, longitude, and altitude as the spatial features, each coordinate is associated with a timestamp. We also round the timestamp to full hours with the assumption that the hours of operation will impact the aircraft's landing time. Additionally, we count the number of aircraft ahead of the target aircraft as the indicators for airspace complexity measure. The airspace complexity largely impacts the workload of the controller, which further leads to potential flight delays due to ATC. 

\subsubsection{Flight Event Recordings  \label{subsubsec: iff-event}}
Flight event recordings are also processed and archived in SDW. IFF flight event data are well-organized tabular format data, instead of time-series coordinates combined with tabular information in flight track data. The timestamp, location, and flight callsign associated with the flight event are stored. \Cref{table: eventdata} shows the detailed descriptions of fifteen different flight event types recorded in the data. We identify three safety-related flight events. Similarly, we process the feature based on the timestamp that the target landing aircraft reaches the defined TMA, which has the same levels (e.g., 10min, 30min, 60min). We count the number of flight events that happened ahead or behind the target aircraft for each level. In such a way, we obtain the flight event recordings as the predicted safety indicators for the target aircraft. \Cref{table: features} lists the name of the flight events processed.

\begin{table}[H]
\caption{List of flight event types in IFF flight event data. The three safety-related flight events are EV\_RRT, EV\_LOOP, and EV\_GOA.}\label{table: eventdata}
\resizebox{\textwidth}{!}{
\begin{tabular}{c|l}
\hline
\rowcolor[HTML]{EFEFEF}
\textbf{Event Type} & \multicolumn{1}{c}{\cellcolor[HTML]{EFEFEF}\textbf{Descriptions}}                                                                                                                             \\ \hline
EV\_TOF             & \begin{tabular}[c]{@{}l@{}}Take off event; defined when the aircraft crosses the\\  departure runway threshold/aircraft wheels come off the pavement.\end{tabular}                          \\ \hline
EV\_USER            & \begin{tabular}[c]{@{}l@{}}User event; a flexible event defined by users preference \\ (e.g., how many times an airplane goes into a center, different volume definitions).\end{tabular}  \\ \hline
EV\_MOF             & \begin{tabular}[c]{@{}l@{}}Mode of flight; a measure of aircraft trajectory in the \\ vertical domain (e.g., descending level, climbing level).\end{tabular}                              \\ \hline
EV\_INIT            & Indicator of flight track beginning recorded by the facility surveillance system.                                                                                                         \\ \hline
EV\_TRNS            & Transition from above/below altitude. Not often used.                                                                                                                                     \\ \hline
EV\_XING            & Crossing event; defined when aircraft is crossing between different spaces.                                                                                                               \\ \hline
\textbf{EV\_RRT}    & \textbf{\begin{tabular}[c]{@{}l@{}}Reroute event; defined when there is a new flight plan issued. \\ Issuing reroutes significantly increases the workload of the ATC, which poses safety concerns.\end{tabular}}                                                                                                                   \\ \hline
EV\_TOC             & Top of climb; defined when the aircraft reaches the cruise altitude.                                                                                                                      \\ \hline
EV\_TOD             & Top of descend; defined when the aircraft starts to descend from the cruise altitude.                                                                                                      \\ \hline
EV\_PXCP            & Unknown event type.                                                                                                                                                                       \\ \hline
EV\_LND             & Landing event; defined when the arrival aircraft passes the arrival runway threshold.                                                                                                     \\ \hline
\textbf{EV\_LOOP}   & \textbf{\begin{tabular}[c]{@{}l@{}}Holding event; defined when the aircraft is circling around in the trajectory. \\ Looping in the TMA increases airspace complexity and leads to safety concerns.\end{tabular}}                                                                                                   \\ \hline
EV\_STOL            & Holding end; defined when the LOOP event ended.                                                                                                                                           \\ \hline
EV\_STOP            & Indicator of stopping flight track recording by the surveillance systems.                                                                                                                     \\ \hline
\textbf{EV\_GOA}    & \textbf{\begin{tabular}[c]{@{}l@{}}Go around event; defined when the aircraft aborted landing in final approach \\ or after touchdown. Go around is a major safety concern.\end{tabular}} \\ \hline
\end{tabular}
}
\end{table}

\subsubsection{Weather Features \label{subsubsec: weather}}
Weather impact is a critical factor of aviation safety and is thus non-negligible in aviation operations. In this work, we also explore performance improvement in machine learning with explicit weather feature inputs. We obtain the wind speed, wind direction, cloud cover, visibility, and humidity near KATL. We use the hourly weather features to record and refer to the full-hour flight monitoring records in \Cref{table: features} to build the final feature table for ML prediction.

\section{Case Study on Scheduling\label{sec: experiments}}

In this section, we introduce machine learning prediction and flight delay optimization case studies. First, the performance evaluation metrics are briefly discussed. Then, we explain the condition-based machine learning predictor for improved performance, where the processed features are classified based on the number of looping event counts. Last, we show that the proposed machine learning-enhanced TSP-TW solution can achieve a shorter total landing time compared to FCFS, for all of the landing aircraft within a time window.

\subsection{Performance Evaluation Metrics}
Proper performance evaluation metrics are required to select the best parameter setup for the machine learning model. Considering a supervised regression problem, we propose three cost functions to address various statistical behaviors. Define a predicted label $y_i$ and the ground truth $\hat{y}_i$, we have,

\noindent \textbf{Mean Absolute Error (MAE)}: MAE is the arithmetic average of the absolute errors between predicted labels and ground truth labels. MAE is a commonly used metric in forecasting and prediction objectives. MAE weighs each sample at the same scale. 
\begin{equation}\label{eq: mae}
    \mathsf{MAE} = \frac{1}{n} \sum_{i=0}^n |y_i - \hat{y_i}|
\end{equation}

\noindent \textbf{Root Mean Squared Error (RMSE)}: RMSE is an alternative to MAE, which share the same drawbacks. RMSE is sensitive to outliers, where a significantly bad prediction aggravates the overall performance measure. This skews the evaluation results towards overestimating the models' badness. 

\begin{equation}\label{eq: rmse}
    \mathsf{RMSE} = \sqrt{\frac{1}{n} \sum_{i=0}^n (y_i - \hat{y_i})^2}
\end{equation}

\noindent \textbf{Root Mean Squared Log Error (RMSLE)}: In ALS predictions, severe delays can happen due to various reasons, which are treated as outliers in data-driven prediction. These outliers are unlikely to be captured by predictors and result in overestimating of model's badness. This can be misleading. We propose to use RMSLE, as shown in \Cref{eq: rmsle}. RMSLE is viewed as the RMSE of log-transformed prediction and log-transformed ground truth. RMSLE is preferred as we need to avoid over-penalizing severe delay scenarios, which helps select the best model parameters.
\begin{equation}\label{eq: rmsle}
    \mathsf{RMSLE} = \sqrt{\frac{1}{n} \sum_{i=0}^n [\log (y_i+1) - \log (\hat{y_i}+1)]^2}
\end{equation}

\subsection{Prediction Analysis}
The model development/training phase has two objectives, predicting the ETA distributions with GBM and formulating the predicted values into optimization for the demonstration of case studies. The first part requires model-tuning efforts. We tackle this from three aspects, 

\begin{table}[H]
    \caption{GBM Hyperparameters and Search Space for Grid Search}
    \label{table: gbm-parameters}
    \begin{tabular}{l|l|l}
    \hline
    \rowcolor[HTML]{EFEFEF} 
    {\color[HTML]{333333} Parameters} & {\color[HTML]{333333} Definition}                                                                           & {\color[HTML]{333333} Search Space} \\ \hline
    learn\_rate                       & The learning rate for the optimizer.                                                                        & {[}0.05, 0.1{]}                     \\ \hline
    max\_depth                        & The maximum tree depth.                                                                                     & {[}7, 8, 9, 10, 11, 12{]}           \\ \hline
    sample\_rate                      & \begin{tabular}[c]{@{}l@{}}The sample without replacement rate\\  along the feature dimension.\end{tabular} & {[}0.8, 1.0{]}                      \\ \hline
    col\_sample\_rate                 & \begin{tabular}[c]{@{}l@{}}The sample without replacement rate\\  along the sample dimension.\end{tabular}  & {[}0.5, 0.6, 0.7, 0.8{]}            \\ \hline
    \end{tabular}
\end{table}

\textbf{Grid Search} is common practice to fine-tune parameterized machine learning predictors, to find the best combination of modifiable hyperparameters. For the GBM used in this work, we are especially interested in the following hyperparameters, a) the learning rate controlling the step size of optimization (efficiency); b) the maximum tree depth to control the order of approximations (accuracy); c) the data sampling rate along the feature and sample dimension (stochasticity). More discussion of b) and c) is in \Cref{subsec: review-gbm}. We list the search space of this study in \Cref{table: gbm-parameters}.

\textbf{Domain Knowledge} and human intelligence can benefit data-driven models. In \Cref{subsec: investigations}, we have discovered the impact of holding patterns on flight delays. The holding pattern is recorded as a looping event in the IFF flight event recordings. Airspace complexity is represented by the number of nearby aircraft of the target landing aircraft. As discussed in \Cref{subsubsec: iff-track} and \Cref{subsubsec: iff-event}, we implicitly include the airspace complexity measurements and flight event number that happened within the certain time range of the target landing aircraft. Lastly, weather features are critical for aviation operations and thus are non-negligible when building ML predictors.

\begin{figure}[h]
    \centering
    \begin{subfigure}[t]{0.485\textwidth}
        \centering
        \includegraphics[width=\textwidth]{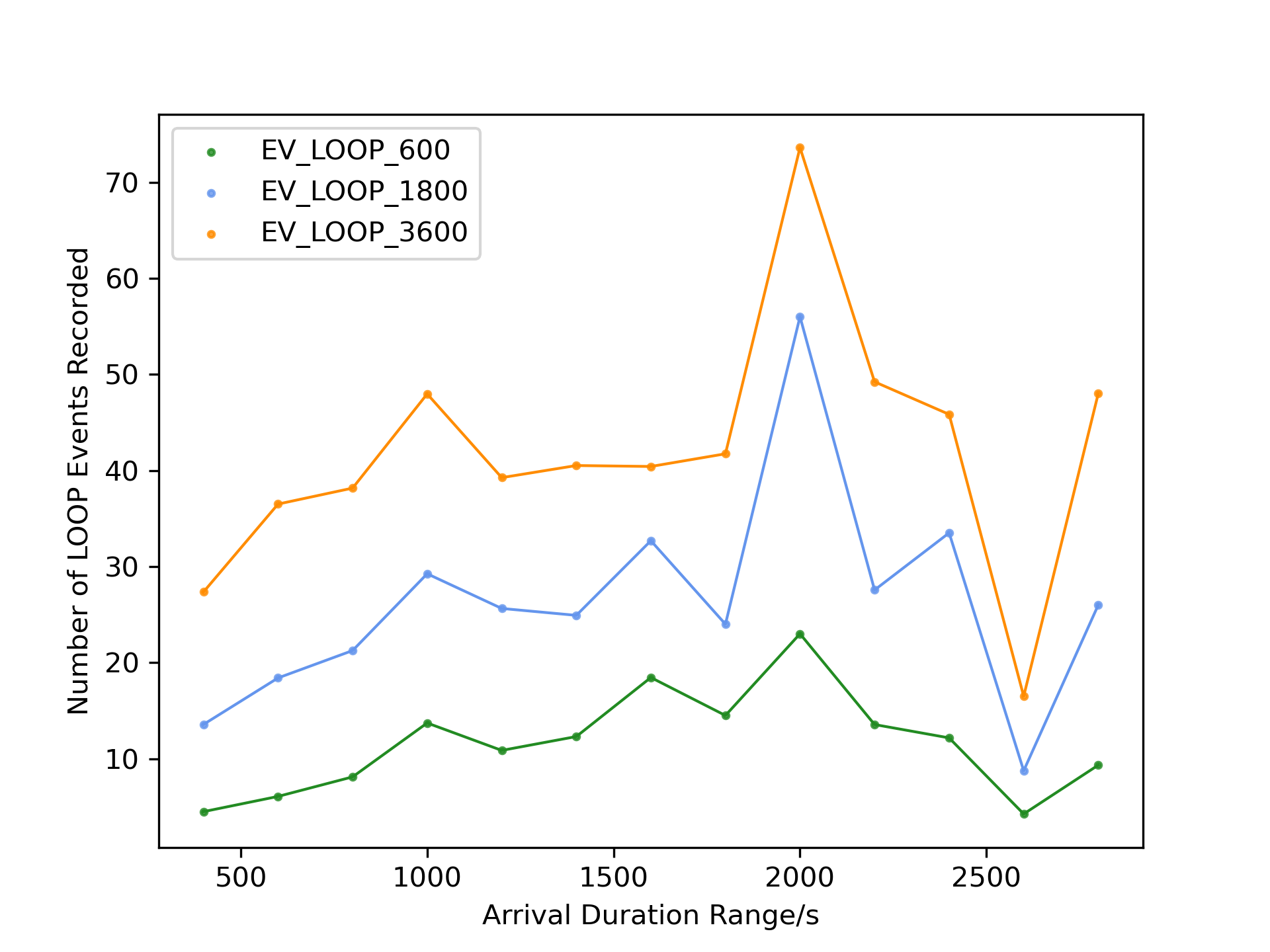}
        \caption{Looping Event Counts v.s. Time Spent from 100NM to 40NM}
        \label{fig: event-hist}
    \end{subfigure}
    ~
    \begin{subfigure}[t]{0.485\textwidth}
        \centering
        \includegraphics[width=\textwidth]{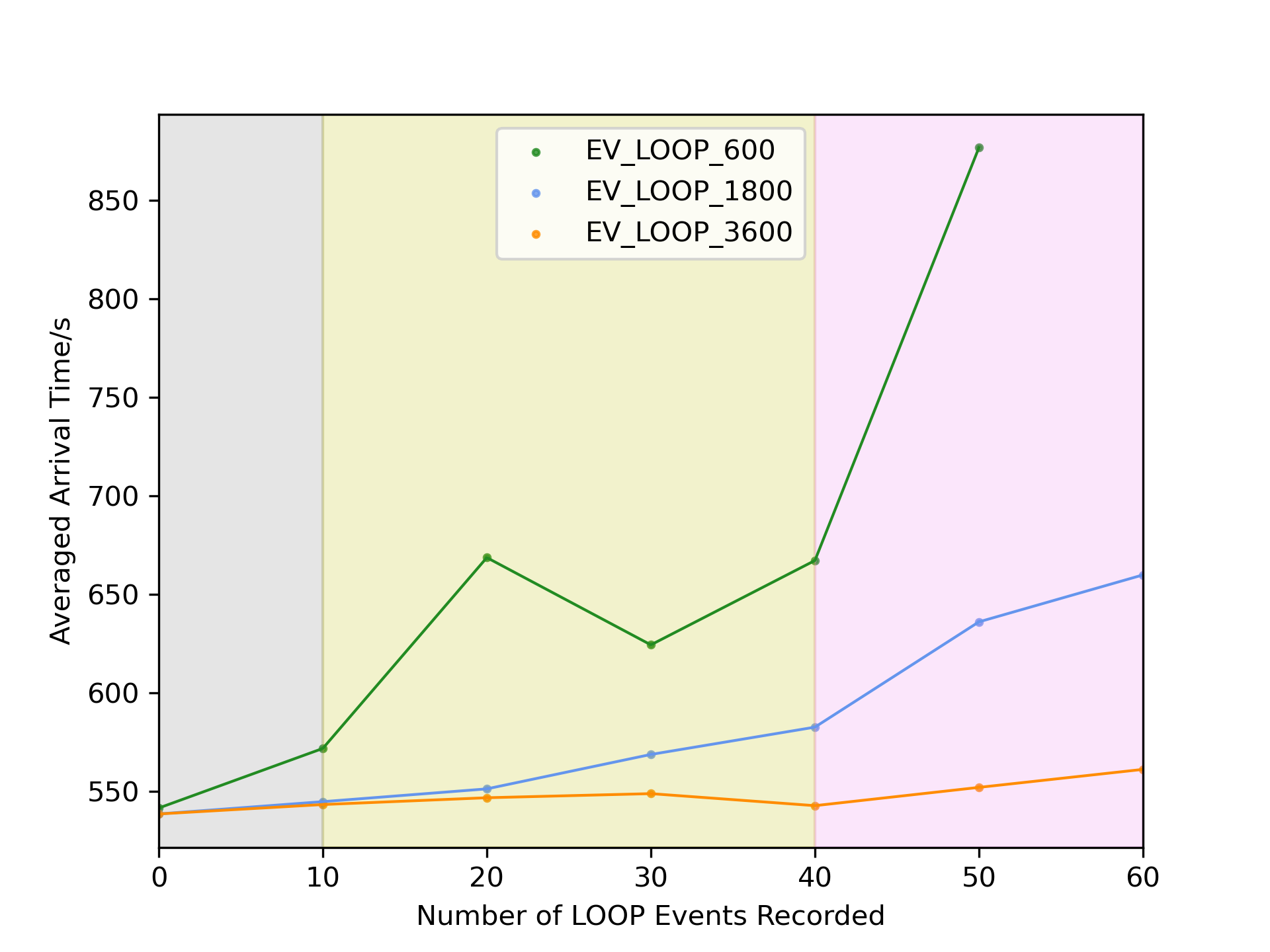}
        \caption{Averaged Time Spent from 100NM to 40NM v.s. Looping Event Counts}
        \label{fig: arr-hist}
    \end{subfigure}
    \caption{Data analysis to identify the reasonable conditions for conditioned prediction. The legend stands for the number of event looping events $600$/$1,800$/$3,600$ seconds ahead/behind the current timestamp.}
    \label{fig: histograms}
\end{figure}

\textbf{Divide-and-Conquer} stands for gaining and maintaining outstanding ML performance divisively. We propose conditional GBM, which pre-filters the data samples based on the number of looping event counts, to gain exceptional prediction capability. \Cref{fig: histograms} shows the statistical analysis between the number of flight events and the arrival aircraft landing time within the $[100, 40]$NM range. \Cref{fig: histograms}(a) shows a non-monotonic trend, where an obvious drop of looping events is presented when the arrival time is greater than $2,500$ seconds. However, from \Cref{fig: histograms}(b), we notice there are approximately three stages along with the increase of looping event counts. We separate the entire dataset into three parts and trained with separate GBM models. We discuss each of the separations as, 

\begin{itemize}
    \item Stage I (EV\_LOOP $\leq$ 10): In this stage, minimum flight event conditions exist, where the arrival time duration is near optimal. At this stage, the arrival time increase is not significant.
    \item Stage II (10 $<$ EV\_LOOP $\leq$ 40): The steady growth stage. At this stage, the arrival time duration is steadily growing with the increase of looping event counts. 
    \item Stage III (EV\_LOOP $>$ 40): The rapidly increasing stage. The arrival time sharply increases when the number of looping events increases. 
\end{itemize}

We collect and process the flight track and flight event data for August 2019 at ARTCC ZTL. A total of $28,181$ well-structured arrival flight information are obtained, with the feature set described in previous sections. Then, the data is filtered for three stages based on the EV\_LOOP\_600 feature. For each stage, we further separate the data into training, validations, and testing set. The training and validation sets are used for GBM training and fine-tuning, and the testing set is used for prediction case studies. 

\begin{figure}[H]
    \centering
    \begin{subfigure}[t]{0.45\textwidth}
        \centering
        \includegraphics[width=\textwidth]{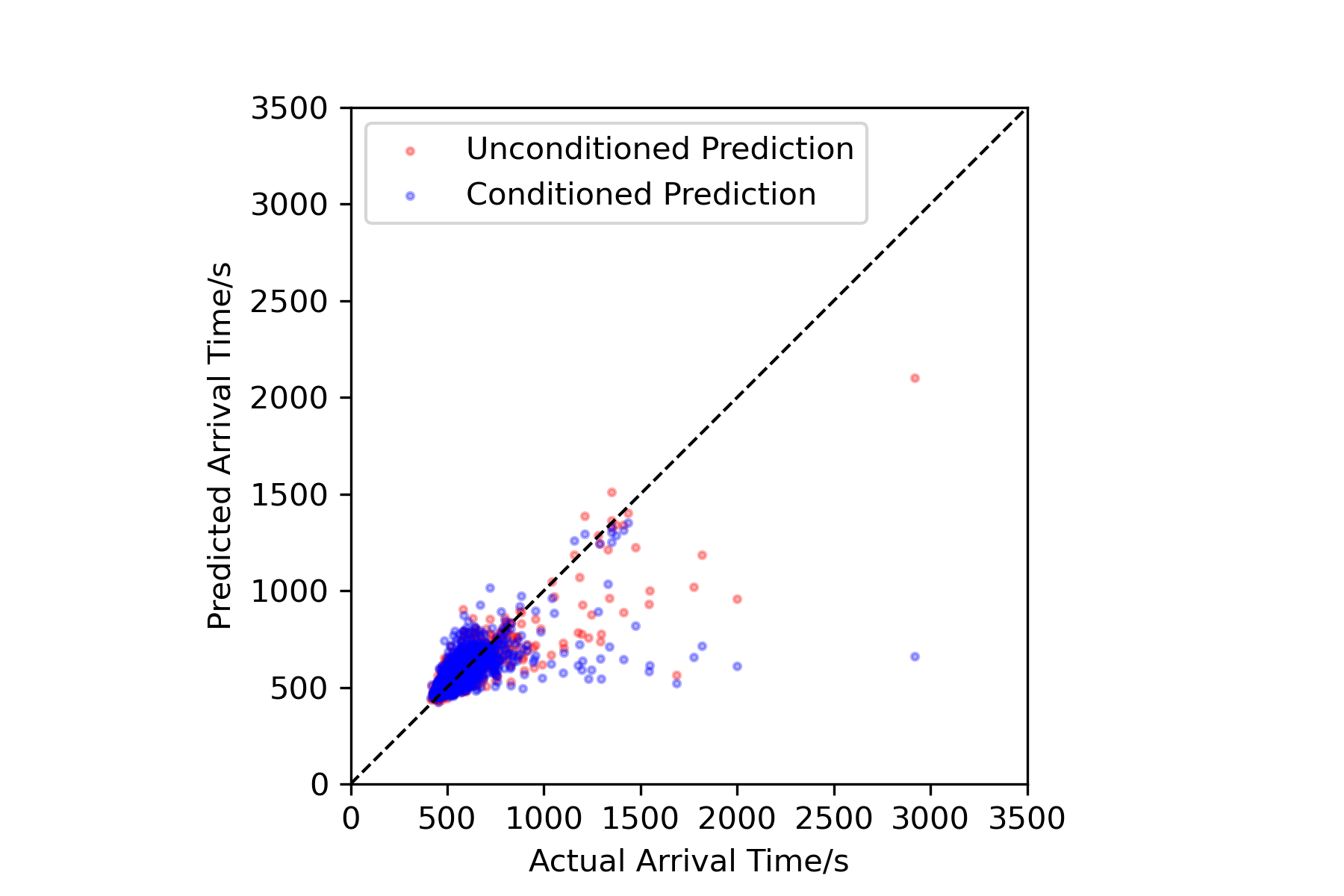}
        \caption{EV\_LOOP $\leq$ 10}
        \label{fig: sel-10}
    \end{subfigure}
    ~
    \begin{subfigure}[t]{0.45\textwidth}
        \centering
        \includegraphics[width=\textwidth]{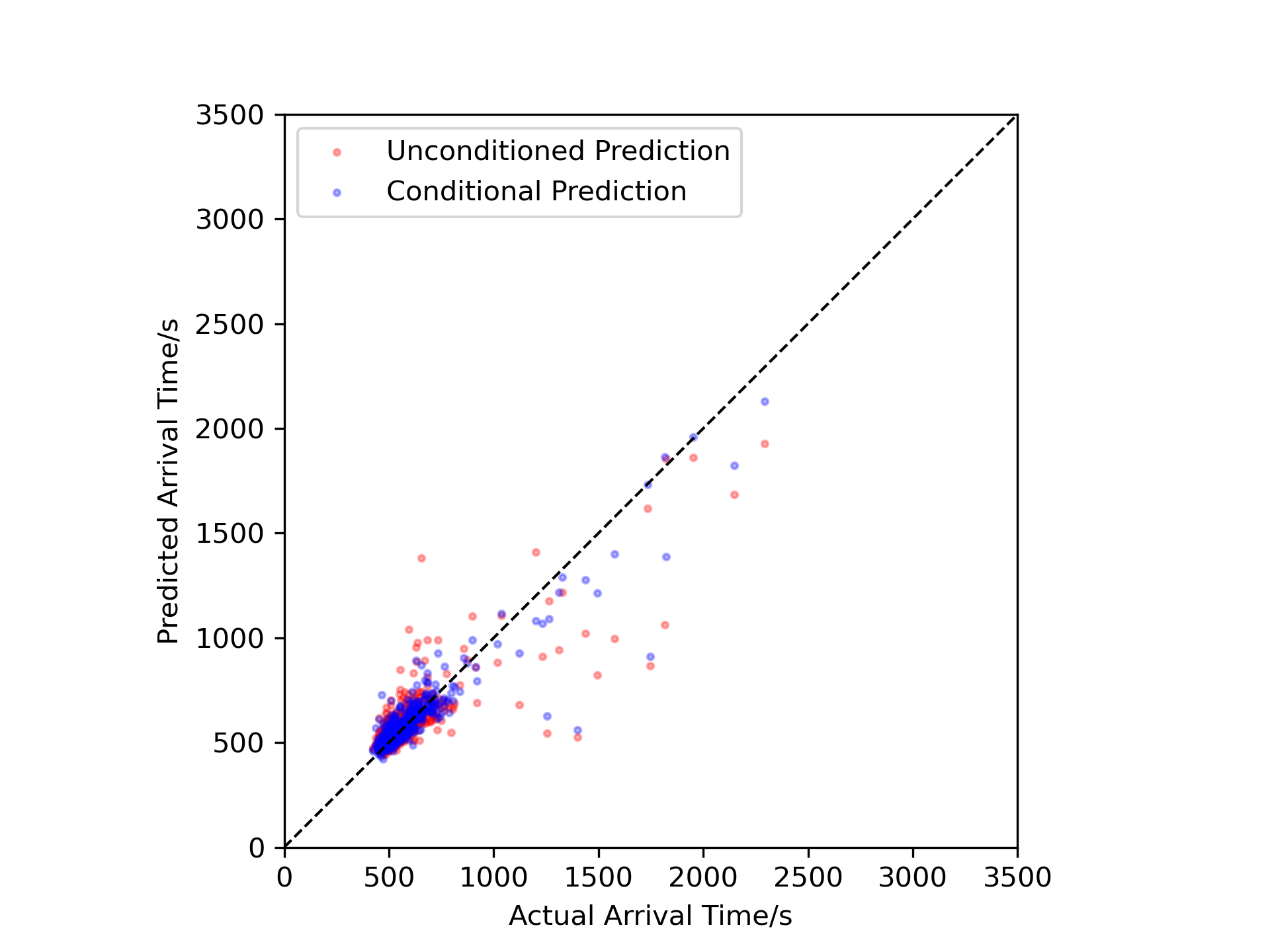}
        \caption{10 $<$ EV\_LOOP $\leq$ 40}
        \label{fig: 10-40}
    \end{subfigure}
    ~
    \begin{subfigure}[t]{0.45\textwidth}
        \centering
        \includegraphics[width=\textwidth]{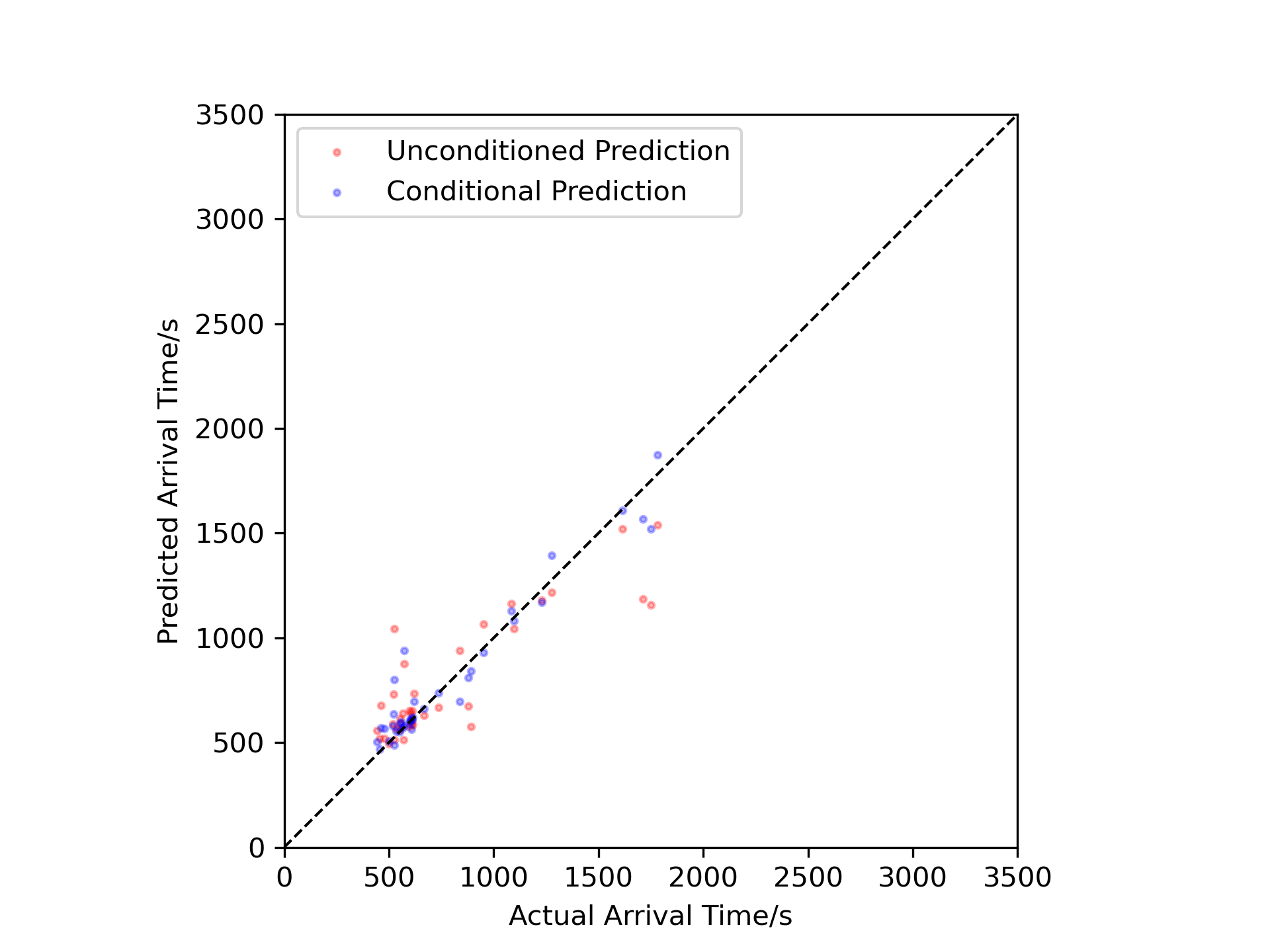}
        \caption{40 $<$ EV\_LOOP}
        \label{fig: 40-60}
    \end{subfigure}
    \caption{Splitting the Testing Set: Unconditioned Prediction v.s. Conditioned Prediction}
    \label{fig: conditionalGBM}
\end{figure}

\begin{figure}[H]
    \centering
    \begin{subfigure}[t]{0.9\textwidth}
        \centering
        \includegraphics[width=\textwidth,height=5cm]{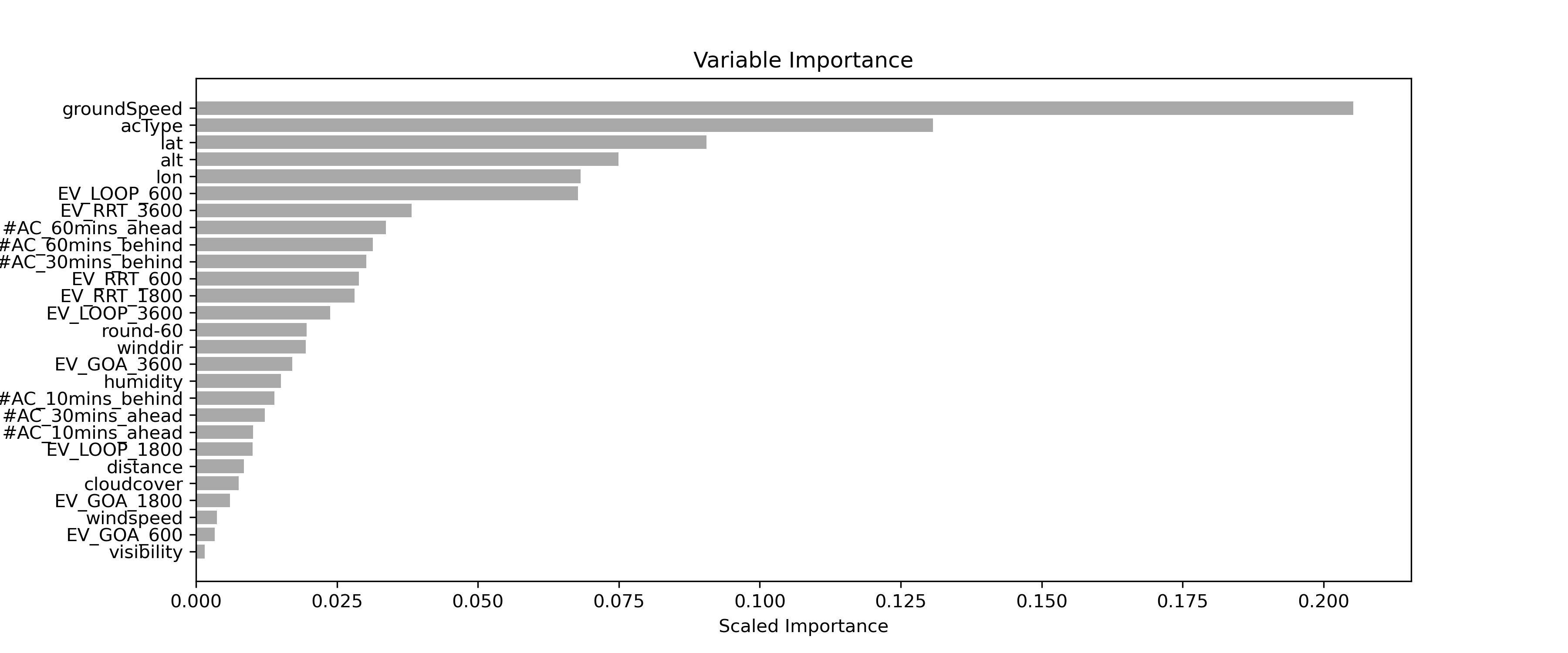}
        \caption{Variable Importance: EV\_LOOP $\leq$ 10}
        \label{fig: vi-sel10}
    \end{subfigure}
    ~
    \begin{subfigure}[t]{0.9\textwidth}
        \centering
        \includegraphics[width=\textwidth,height=5cm]{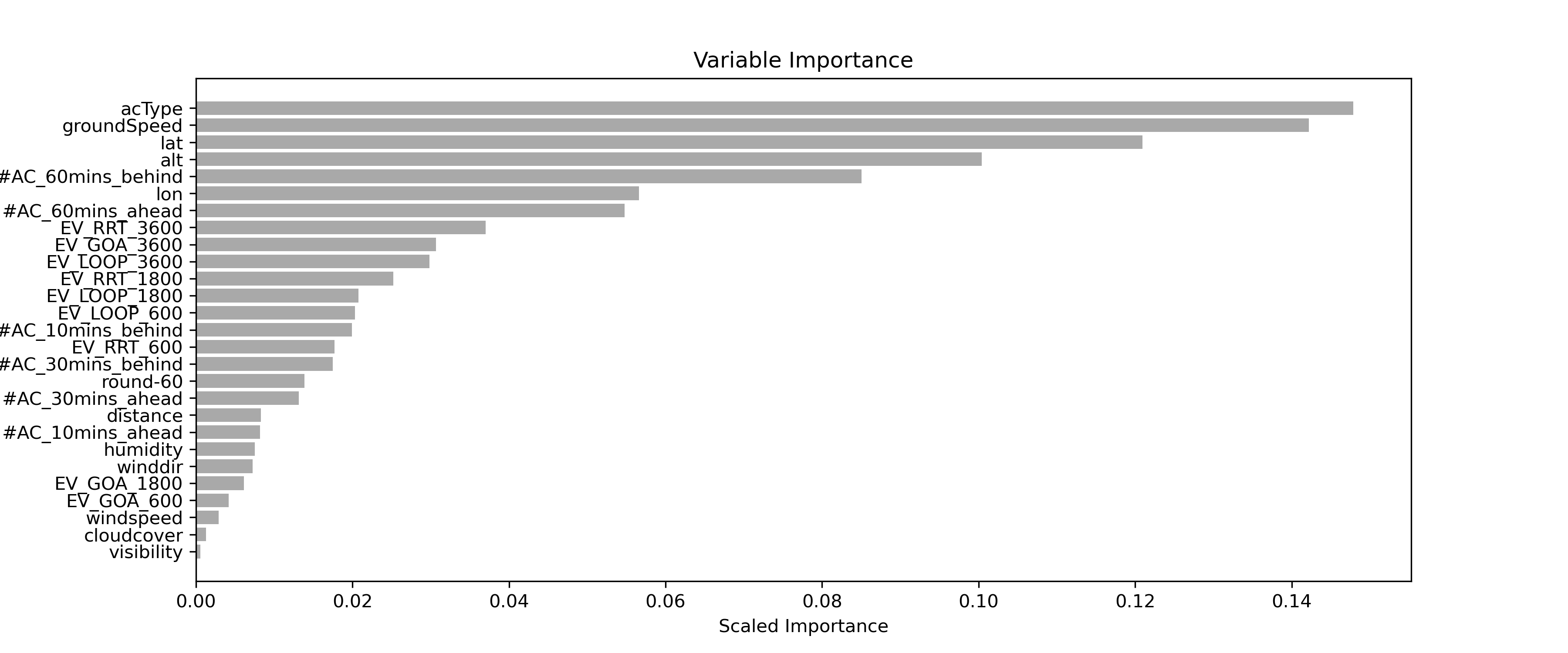}
        \caption{Variable Importance: 10 $<$ EV\_LOOP $\leq$ 40}
        \label{fig: vi-10-40}
    \end{subfigure}
    ~
    \begin{subfigure}[t]{0.9\textwidth}
        \centering
        \includegraphics[width=\textwidth,height=5cm]{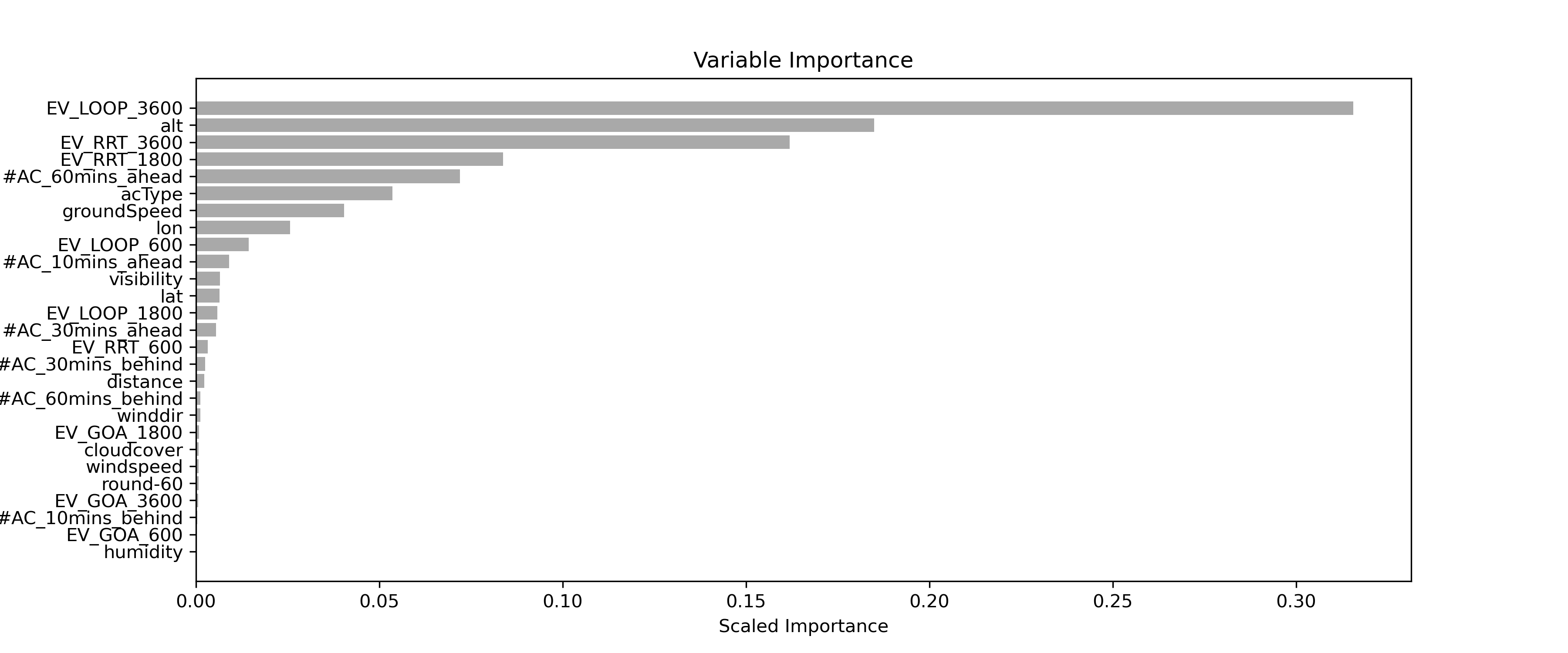}
        \caption{Variable Importance: 40 $<$ EV\_LOOP}
        \label{fig: vi-40-60}
    \end{subfigure}
    \caption{Variable importance for three divisive predictors. The importance of the speed profile keeps decreasing with the increase in the number of looping events. In (c), the dominant variables are airspace complexity, and safety-related event counts.}
    \label{fig: vi-conditioinalGBM}
\end{figure}

In \Cref{table: gbm-results}, we evaluate the performance of the three-stage model using the testing dataset and compare it with the unconditional method without considering different growth behaviors. Including the flight event-related features and weather features boost the model performance by a large margin, while the conditioned predictor further refines the results. \Cref{fig: conditionalGBM} shows the evaluation results in comparison for unconditioned predictions and conditioned predictions. \Cref{fig: vi-conditioinalGBM} shows the corresponding variable (feature) importance for conditioned predictors. For stage I, the conditioned prediction doesn't significantly improve the prediction accuracy. At this stage, the number of looping events is not a critical variable for the GBM. The ground speed is prominently dominant in the stage I predictions. However, for stage II and stage III, the conditioned predictor greatly enhanced the performance. At stage II, the type of aircraft shows nearly the same variable importance as the ground speed. Notably, at stage III, the ground speed turns into a less critical factor contributing to the overall model performance. The airspace complexity factors, geological information, weather features, and flight safety-related events are essential.

\begin{table}[H]
    \caption{GBM Evaluation Results on the Testing Dataset (in seconds)}
    \label{table: gbm-results}
    \begin{tabular}{l|l|l|l}
    \hline
    \rowcolor[HTML]{EFEFEF} 
    Model                                       & RMSE    & MAE    & RMSLE \\ \hline
    Predictor                                   & 118.147 & 42.139 & 1.136 \\ \hline
    Predictor w/ Events                         & 93.784  & 38.085 & 1.011 \\ \hline
    Predictor w/ Weather                        & 98.872  & 37.766 & 0.099 \\ \hline
    Predictor w/ Events and Weather             & 83.022  & 36.739 & 0.096 \\ \hline
    Conditioned Predictor w/ Events and Weather & 78.181  & 35.690 & 0.096 \\ \hline
    \end{tabular}
\end{table}

\subsection{ALS Case Studies}
\begin{figure}[H]
    \centering
    \includegraphics[width=0.85\textwidth]{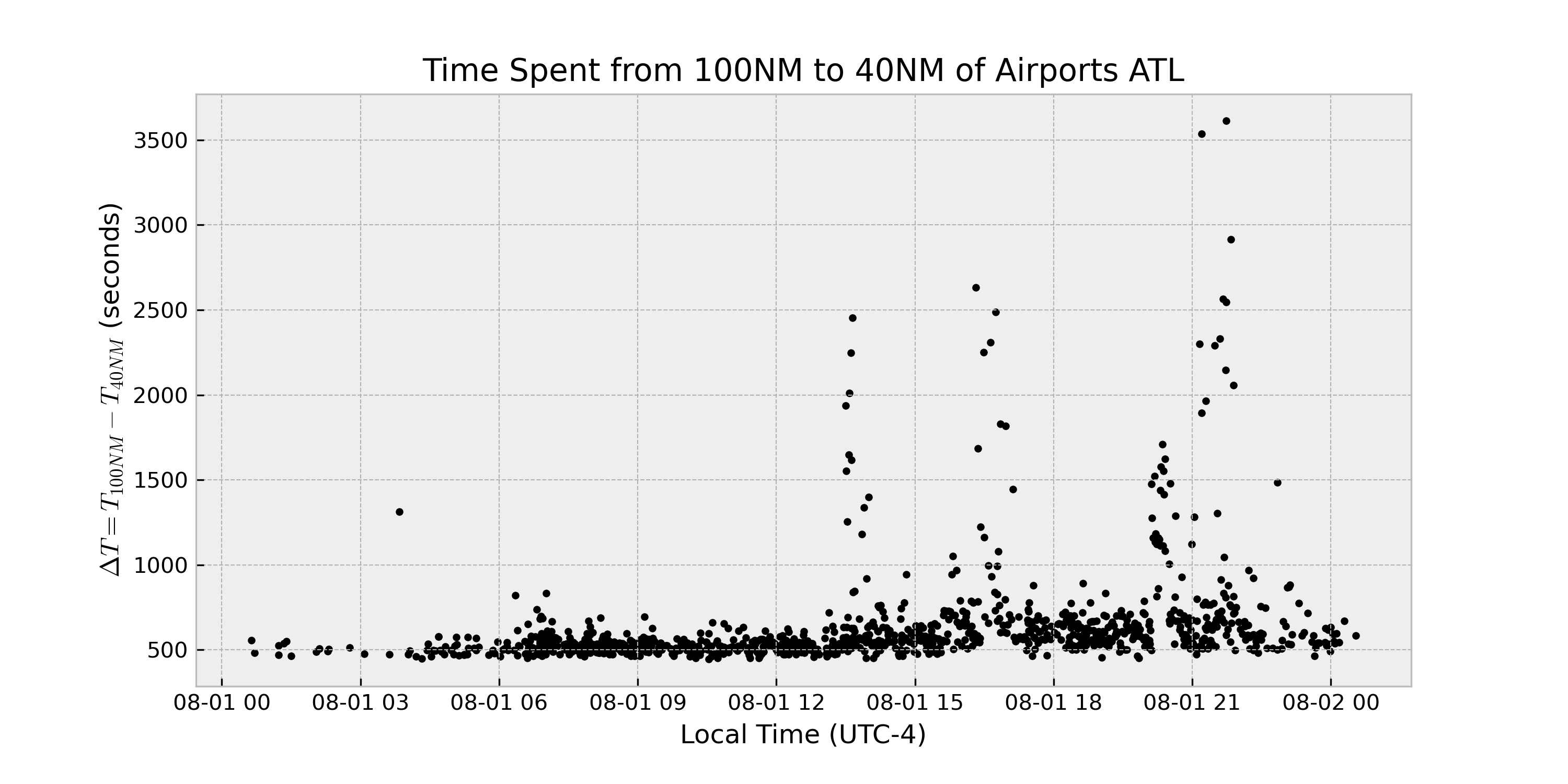}
    \caption{Scatter plot of flight times around KATL TMA on Aug 1st, 2019. This figure shows that there are three severe delay periods, starting from around 13:20, 16:00, 21:00, respectively.}
    \label{fig: arrival-0801}
\end{figure}

\begin{figure}[H]
    \centering
    \begin{subfigure}[t]{0.45\textwidth}
        \centering
        \includegraphics[width=\textwidth]{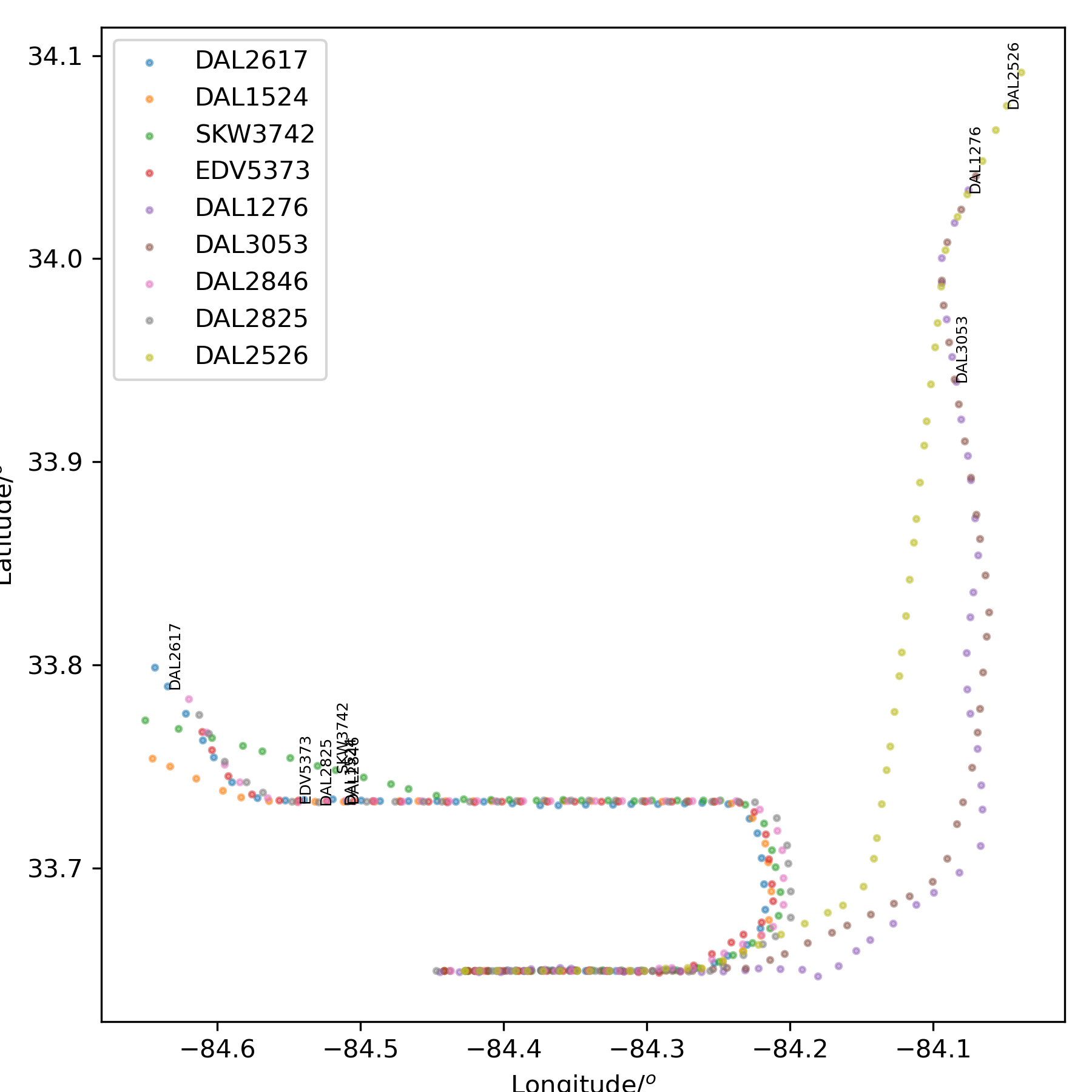}
        \caption{Case Study I: Landing Trajectories on KATL Runway 26R}
        \label{fig: case1-traj}
    \end{subfigure}
    ~
    \begin{subfigure}[t]{0.45\textwidth}
        \centering
        \includegraphics[width=\textwidth]{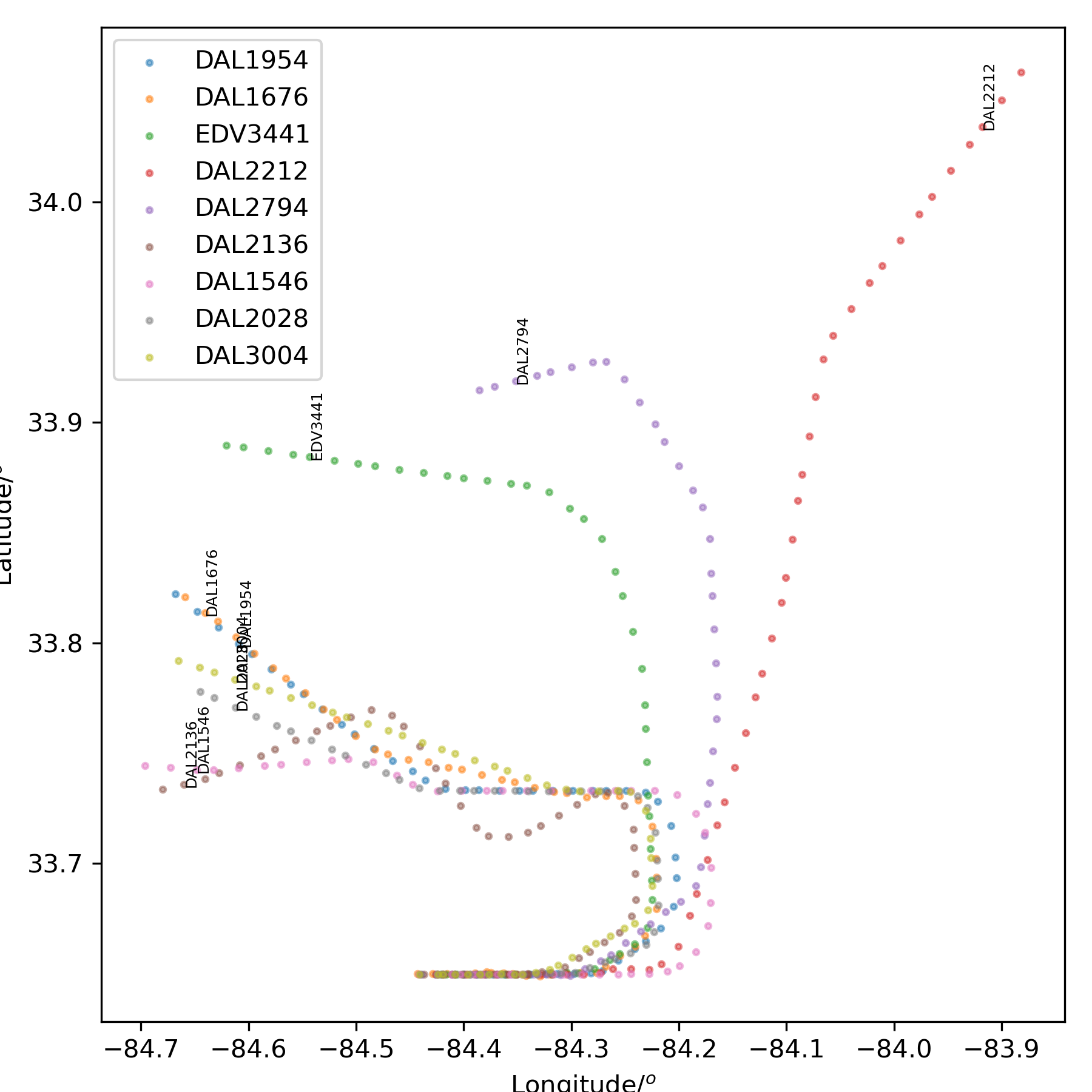}
        \caption{Case Study II: Landing Trajectories on KATL Runway 26R}
        \label{fig: case2-traj}
    \end{subfigure}
    ~
    \begin{subfigure}[t]{0.45\textwidth}
        \centering
        \includegraphics[width=\textwidth]{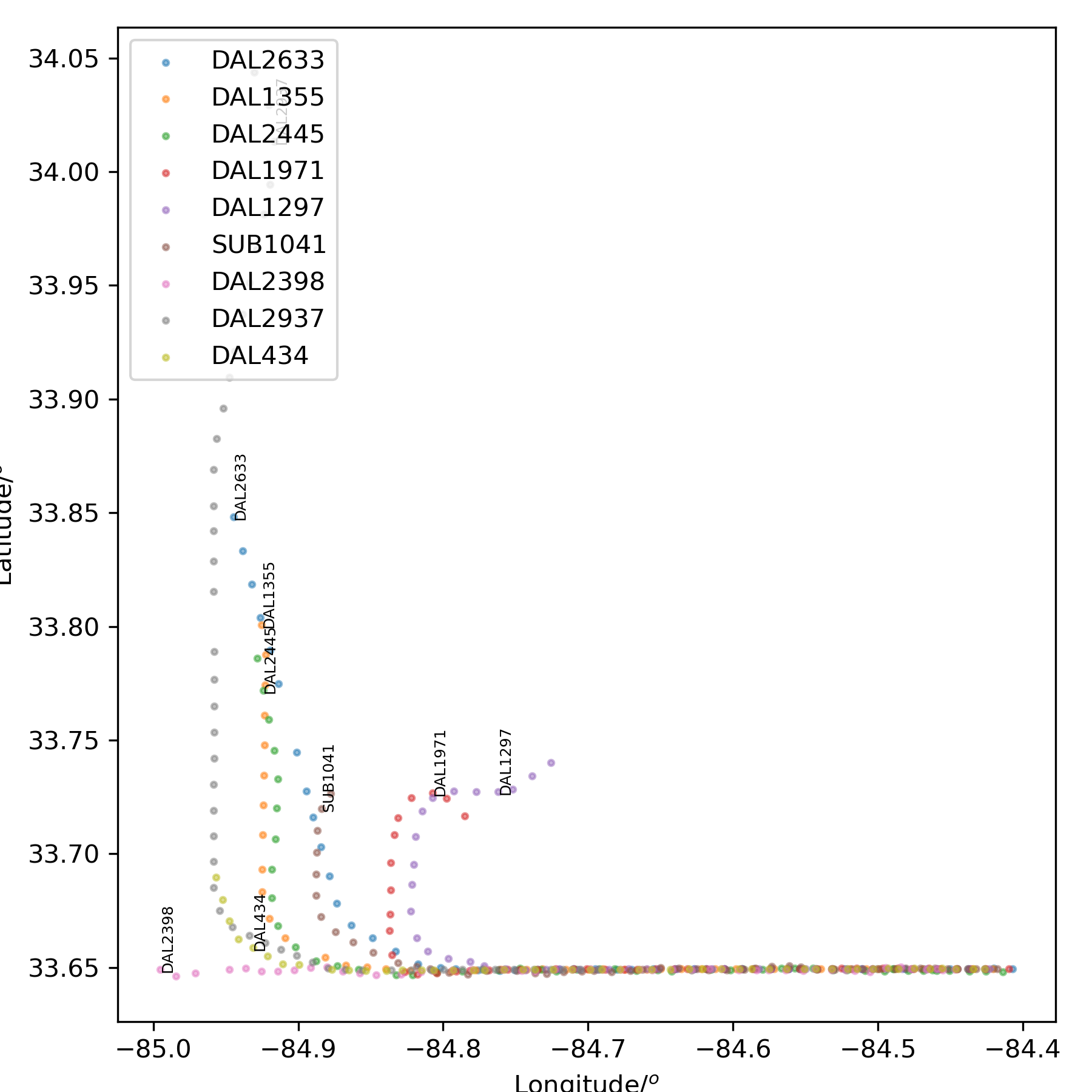}
        \caption{Case Study III: Landing Trajectories on KATL Runway 26R}
        \label{fig: case3-traj}
    \end{subfigure}
    \caption{Visualization of landing trajectories for two case studies on KATL Runway 26R: a) 13:20 to 13:45, Aug 1st, 2019; b) 16:00 - 16:30, Aug 1st, 2019; c) 21:00 - 21:20, Aug 1st, 2019.}
    \label{fig: results-cases-traj}
\end{figure}

In this section, we discuss real-world demonstration case studies. At first, we define the problem horizon by visualizing and analyzing the real-world data. \Cref{fig: arrival-0801} shows the scatter plot of time spent for approaching flights around KATL TML on Aug 1st, 2019. The increased scatter density indicates the business of aviation operations at a certain timestamp. The Y-axis denotes the flight time from $100$NM to $40$NM away from the terminal TMA. It's clear that the normal time spent should be lower than $1,000$s, which are also cross-referenced by \Cref{fig: boxplots}(a) and \Cref{fig: delay-compare}(b). Based on these, we conclude that there are three severely delayed time ranges, starting from 13:00, 16:00, and 21:00, approximately. In this demonstration, we present three case studies. In the first scenario, we take 9 successive landing flights from 13:20 to 13:45 and show the trajectories in \Cref{fig: results-cases-traj}(a). These landing flights mostly follow two groups of landing procedures. The first group of 6 flights came from the west side of KATL for a west landing on KATL runway 26R, while another group of landing aircraft came from the northeast for the same west landing at 26R. Similarly, in the second case study, we set the time duration starting from 16:00 to 16:30. Case II has three groups of landing procedures, where flight EDV3441 and DAL2794 maneuvered for west landing ahead of time. Case III considers all west landing scenarios within the 200NM TMA. The selection of time windows is based on the availability of aircraft landing on the same runway, and the scalability of the TSP-TW solver.

From GBM, we obtain the distributions for successive landing aircraft pair $t_i \sim \mathcal{N}(\mu_i, \sigma_i)$, $t_j \sim \mathcal{N}(\mu_j, \sigma_j)$. To determine the $\mathcal{T}_{ij}$ for various aircraft types, we refer to FAA Order JO 7360.1H \citep{faa-order7360.1}. Following \Cref{eq: t-t-re}, we obtain the distribution of $t_{ij}$. Given the separation violation probability, we can estimate the probability intervals with numerical tools. In this way, we obtain $\Check{\Check{t_{ij}}}$, $u_i$, and $l_i$ from the learning of historical data. Then, we incorporate the learned parameters into the formulation of TSP-TW and solve with Python optimization solvers \citep{hart2017pyomo}. It's worth pointing out that, the aircraft involved in our case studies are classified as medium size aircraft. Based on \Cref{table: mst-table}, we choose the \textit{Large-Large} minimum separation threshold, $64$ seconds, for our case studies. We set the number of landing aircraft to 9 for both cases due to the computational complexity, which corresponds to at least $\sim10$min optimization horizon in extreme scenarios. The exploration of other efficient solvers for MILP is beyond the focus of this study.

\begin{table}[H]
\caption{The detailed ALS results for three case studies. The determination of the case study windows has been discussed. Depending on FAA Order JO 7360.1H \citep{faa-order7360.1}, all of the aircraft involved in these case studies belong to the medium-weight class. Referring to \Cref{table: mst-table}, we use the Large-Large $\mathcal{T}_{ij}$ to be fixed at 64. All of the Unix timestamps contained in the data have been transformed to the local time zone at ARTCC ZTL (UTC-4).}
\label{table: case-study-results}
\centering
\resizebox{0.75\textwidth}{!}
{  
\begin{tabular}{ccccccc}
\hline
\multicolumn{7}{c}{\cellcolor[HTML]{C0C0C0}Case Study I: 13:20 - 13:45}                                                                                                                                                                                                                                                                                                        \\ \hline
\multicolumn{1}{c|}{\begin{tabular}[c]{@{}c@{}}Entering\\ Sequence\end{tabular}} & \multicolumn{1}{c|}{Callsign} & \multicolumn{1}{c|}{AcType} & \multicolumn{1}{c|}{$\mathcal{T}_{ij}$/s} & \multicolumn{1}{c|}{\begin{tabular}[c]{@{}c@{}}Landing\\ Sequence\end{tabular}} & \multicolumn{1}{c|}{\begin{tabular}[c]{@{}c@{}}Landing\\ Time\end{tabular}} & $\sigma_{ij}$/s \\ \hline
\multicolumn{1}{c|}{1}                                                           & \multicolumn{1}{c|}{DAL2617}  & \multicolumn{1}{c|}{B712}   & \multicolumn{1}{c|}{64}                   & \multicolumn{1}{c|}{3}                                                          & \multicolumn{1}{c|}{13:48:32}                                               & 4.52            \\ \hline
\multicolumn{1}{c|}{2}                                                           & \multicolumn{1}{c|}{DAL1524}  & \multicolumn{1}{c|}{B752}   & \multicolumn{1}{c|}{64}                   & \multicolumn{1}{c|}{4}                                                          & \multicolumn{1}{c|}{13:49:45}                                               & 6.35            \\ \hline
\multicolumn{1}{c|}{3}                                                           & \multicolumn{1}{c|}{SKW3742}  & \multicolumn{1}{c|}{CRJ2}   & \multicolumn{1}{c|}{64}                   & \multicolumn{1}{c|}{1}                                                          & \multicolumn{1}{c|}{13:44:47}                                               & 5.15            \\ \hline
\multicolumn{1}{c|}{4}                                                           & \multicolumn{1}{c|}{EDV5373}  & \multicolumn{1}{c|}{CRJ7}   & \multicolumn{1}{c|}{64}                   & \multicolumn{1}{c|}{7}                                                          & \multicolumn{1}{c|}{13:53:28}                                               & 0.15            \\ \hline
\multicolumn{1}{c|}{5}                                                           & \multicolumn{1}{c|}{DAL1276}  & \multicolumn{1}{c|}{MD88}   & \multicolumn{1}{c|}{64}                   & \multicolumn{1}{c|}{5}                                                          & \multicolumn{1}{c|}{13:50:56}                                               & 3.00            \\ \hline
\multicolumn{1}{c|}{6}                                                           & \multicolumn{1}{c|}{DAL3053}  & \multicolumn{1}{c|}{MD88}   & \multicolumn{1}{c|}{64}                   & \multicolumn{1}{c|}{2}                                                          & \multicolumn{1}{c|}{13:45:58}                                               & 2.52            \\ \hline
\multicolumn{1}{c|}{7}                                                           & \multicolumn{1}{c|}{DAL2846}  & \multicolumn{1}{c|}{MD88}   & \multicolumn{1}{c|}{64}                   & \multicolumn{1}{c|}{6}                                                          & \multicolumn{1}{c|}{13:52:09}                                               & 5.30            \\ \hline
\multicolumn{1}{c|}{8}                                                           & \multicolumn{1}{c|}{DAL2825}  & \multicolumn{1}{c|}{B738}   & \multicolumn{1}{c|}{64}                   & \multicolumn{1}{c|}{8}                                                          & \multicolumn{1}{c|}{13:54:59}                                               & 21.24           \\ \hline
\multicolumn{1}{c|}{9}                                                           & \multicolumn{1}{c|}{DAL2526}  & \multicolumn{1}{c|}{MD88}   & \multicolumn{1}{c|}{64}                   & \multicolumn{1}{c|}{9}                                                          & \multicolumn{1}{c|}{13:57:17}                                               & 8.17            \\ \hline
\multicolumn{7}{c}{\cellcolor[HTML]{C0C0C0}Case Study II: 16:00 - 16:30}                                                                                                                                                                                                                                                                                                     \\ \hline
\multicolumn{1}{c|}{\begin{tabular}[c]{@{}c@{}}Entering\\ Sequence\end{tabular}} & \multicolumn{1}{c|}{Callsign} & \multicolumn{1}{c|}{AcType} & \multicolumn{1}{c|}{$\mathcal{T}_{ij}$/s} & \multicolumn{1}{c|}{\begin{tabular}[c]{@{}c@{}}Landing\\ Sequence\end{tabular}} & \multicolumn{1}{c|}{\begin{tabular}[c]{@{}c@{}}Landing\\ Time\end{tabular}} & $\sigma_{ij}$/s \\ \hline
\multicolumn{1}{c|}{1}                                                           & \multicolumn{1}{c|}{DAL1954}  & \multicolumn{1}{c|}{A321}   & \multicolumn{1}{c|}{64}                   & \multicolumn{1}{c|}{2}                                                          & \multicolumn{1}{c|}{16:30:46}                                               & 26.42           \\ \hline
\multicolumn{1}{c|}{2}                                                           & \multicolumn{1}{c|}{DAL1676}  & \multicolumn{1}{c|}{MD88}   & \multicolumn{1}{c|}{64}                   & \multicolumn{1}{c|}{1}                                                          & \multicolumn{1}{c|}{16:28:45}                                               & 35.36           \\ \hline
\multicolumn{1}{c|}{3}                                                           & \multicolumn{1}{c|}{EDV3441}  & \multicolumn{1}{c|}{CRJ9}   & \multicolumn{1}{c|}{64}                   & \multicolumn{1}{c|}{3}                                                          & \multicolumn{1}{c|}{16:41:59}                                               & 4.88            \\ \hline
\multicolumn{1}{c|}{4}                                                           & \multicolumn{1}{c|}{DAL2212}  & \multicolumn{1}{c|}{B712}   & \multicolumn{1}{c|}{64}                   & \multicolumn{1}{c|}{5}                                                          & \multicolumn{1}{c|}{17:02:53}                                               & 16.63           \\ \hline
\multicolumn{1}{c|}{5}                                                           & \multicolumn{1}{c|}{DAL2794}  & \multicolumn{1}{c|}{B739}   & \multicolumn{1}{c|}{64}                   & \multicolumn{1}{c|}{6}                                                          & \multicolumn{1}{c|}{17:11:57}                                               & 27.68           \\ \hline
\multicolumn{1}{c|}{6}                                                           & \multicolumn{1}{c|}{DAL2136}  & \multicolumn{1}{c|}{MD90}   & \multicolumn{1}{c|}{64}                   & \multicolumn{1}{c|}{8}                                                          & \multicolumn{1}{c|}{17:21:53}                                               & 104.62          \\ \hline
\multicolumn{1}{c|}{7}                                                           & \multicolumn{1}{c|}{DAL1546}  & \multicolumn{1}{c|}{A321}   & \multicolumn{1}{c|}{64}                   & \multicolumn{1}{c|}{7}                                                          & \multicolumn{1}{c|}{17:17:42}                                               & 145.11          \\ \hline
\multicolumn{1}{c|}{8}                                                           & \multicolumn{1}{c|}{DAL2028}  & \multicolumn{1}{c|}{A321}   & \multicolumn{1}{c|}{64}                   & \multicolumn{1}{c|}{4}                                                          & \multicolumn{1}{c|}{16:43:40}                                               & 163.95          \\ \hline
\multicolumn{1}{c|}{9}                                                           & \multicolumn{1}{c|}{DAL3004}  & \multicolumn{1}{c|}{B739}   & \multicolumn{1}{c|}{64}                   & \multicolumn{1}{c|}{9}                                                          & \multicolumn{1}{c|}{17:26:21}                                               & 158.20          \\ \hline
\multicolumn{7}{c}{\cellcolor[HTML]{C0C0C0}Case Study III: 21:00 - 21:20}                                                                                                                                                                                                                                                                                                    \\ \hline
\multicolumn{1}{c|}{\begin{tabular}[c]{@{}c@{}}Entering\\ Sequence\end{tabular}} & \multicolumn{1}{c|}{Callsign} & \multicolumn{1}{c|}{AcType} & \multicolumn{1}{c|}{$\mathcal{T}_{ij}$/s} & \multicolumn{1}{c|}{\begin{tabular}[c]{@{}c@{}}Landing\\ Sequence\end{tabular}} & \multicolumn{1}{c|}{\begin{tabular}[c]{@{}c@{}}Landing\\ Time\end{tabular}} & $\sigma_{ij}$/s \\ \hline
\multicolumn{1}{c|}{1}                                                           & \multicolumn{1}{c|}{DAL2633}  & \multicolumn{1}{c|}{B739}   & \multicolumn{1}{c|}{64}                   & \multicolumn{1}{c|}{4}                                                          & \multicolumn{1}{c|}{21:43:07}                                               & 25.72           \\ \hline
\multicolumn{1}{c|}{2}                                                           & \multicolumn{1}{c|}{DAL1355}  & \multicolumn{1}{c|}{B739}   & \multicolumn{1}{c|}{64}                   & \multicolumn{1}{c|}{3}                                                          & \multicolumn{1}{c|}{21:41:29}                                               & 76.47           \\ \hline
\multicolumn{1}{c|}{3}                                                           & \multicolumn{1}{c|}{DAL2445}  & \multicolumn{1}{c|}{B712}   & \multicolumn{1}{c|}{64}                   & \multicolumn{1}{c|}{5}                                                          & \multicolumn{1}{c|}{21:44:45}                                               & 2.03            \\ \hline
\multicolumn{1}{c|}{4}                                                           & \multicolumn{1}{c|}{DAL1971}  & \multicolumn{1}{c|}{MD88}   & \multicolumn{1}{c|}{64}                   & \multicolumn{1}{c|}{1}                                                          & \multicolumn{1}{c|}{21:38:12}                                               & 18.37           \\ \hline
\multicolumn{1}{c|}{5}                                                           & \multicolumn{1}{c|}{DAL1297}  & \multicolumn{1}{c|}{MD88}   & \multicolumn{1}{c|}{64}                   & \multicolumn{1}{c|}{7}                                                          & \multicolumn{1}{c|}{21:54:14}                                               & 7.54            \\ \hline
\multicolumn{1}{c|}{6}                                                           & \multicolumn{1}{c|}{SUB1041}  & \multicolumn{1}{c|}{B190}   & \multicolumn{1}{c|}{64}                   & \multicolumn{1}{c|}{8}                                                          & \multicolumn{1}{c|}{21:56:57}                                               & 9.69            \\ \hline
\multicolumn{1}{c|}{7}                                                           & \multicolumn{1}{c|}{DAL2398}  & \multicolumn{1}{c|}{B712}   & \multicolumn{1}{c|}{64}                   & \multicolumn{1}{c|}{6}                                                          & \multicolumn{1}{c|}{21:50:28}                                               & 5.69            \\ \hline
\multicolumn{1}{c|}{8}                                                           & \multicolumn{1}{c|}{DAL2937}  & \multicolumn{1}{c|}{A321}   & \multicolumn{1}{c|}{64}                   & \multicolumn{1}{c|}{2}                                                          & \multicolumn{1}{c|}{21:39:54}                                               & 125.96          \\ \hline
\multicolumn{1}{c|}{9}                                                           & \multicolumn{1}{c|}{DAL434}   & \multicolumn{1}{c|}{MD90}   & \multicolumn{1}{c|}{64}                   & \multicolumn{1}{c|}{9}                                                          & \multicolumn{1}{c|}{21:59:42}                                               & 19.57           \\ \hline
\end{tabular}
}

\end{table}

\begin{figure}[H]
    \centering
    \begin{subfigure}[t]{0.95\textwidth}
        \centering
        \includegraphics[width=\textwidth]{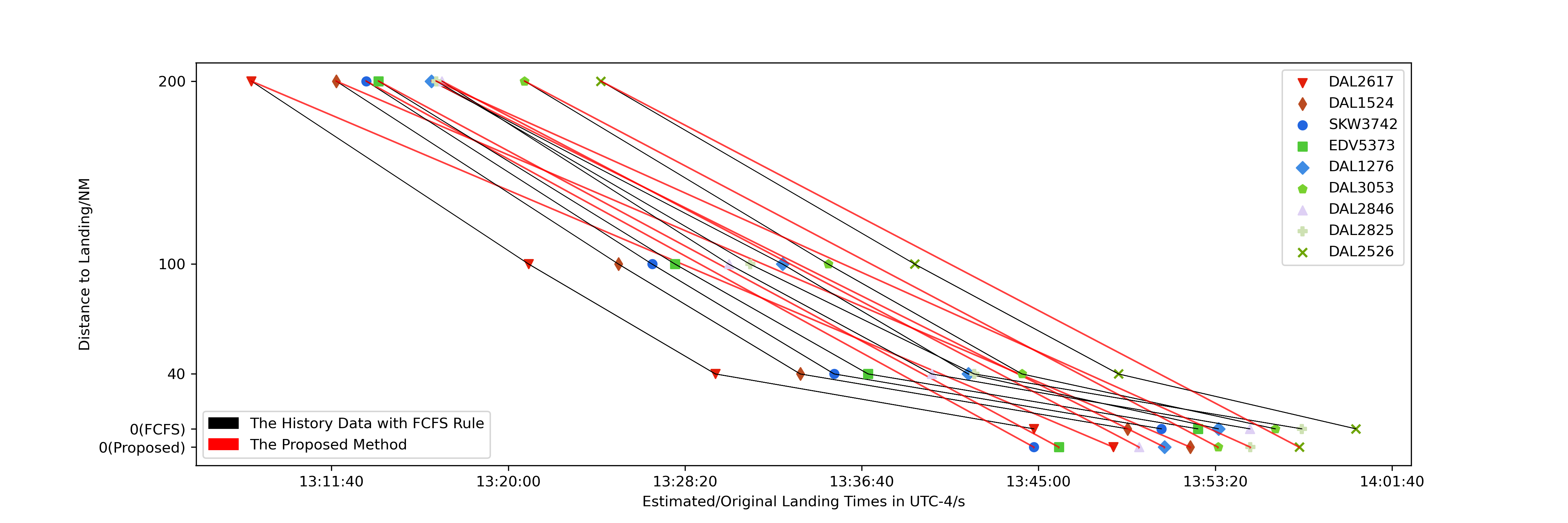}
        \caption{Case Study I: Proposed Method v.s. FCFS}
        \label{fig: results-case1}
    \end{subfigure}
    ~
    \begin{subfigure}[t]{0.95\textwidth}
        \centering
        \includegraphics[width=\textwidth]{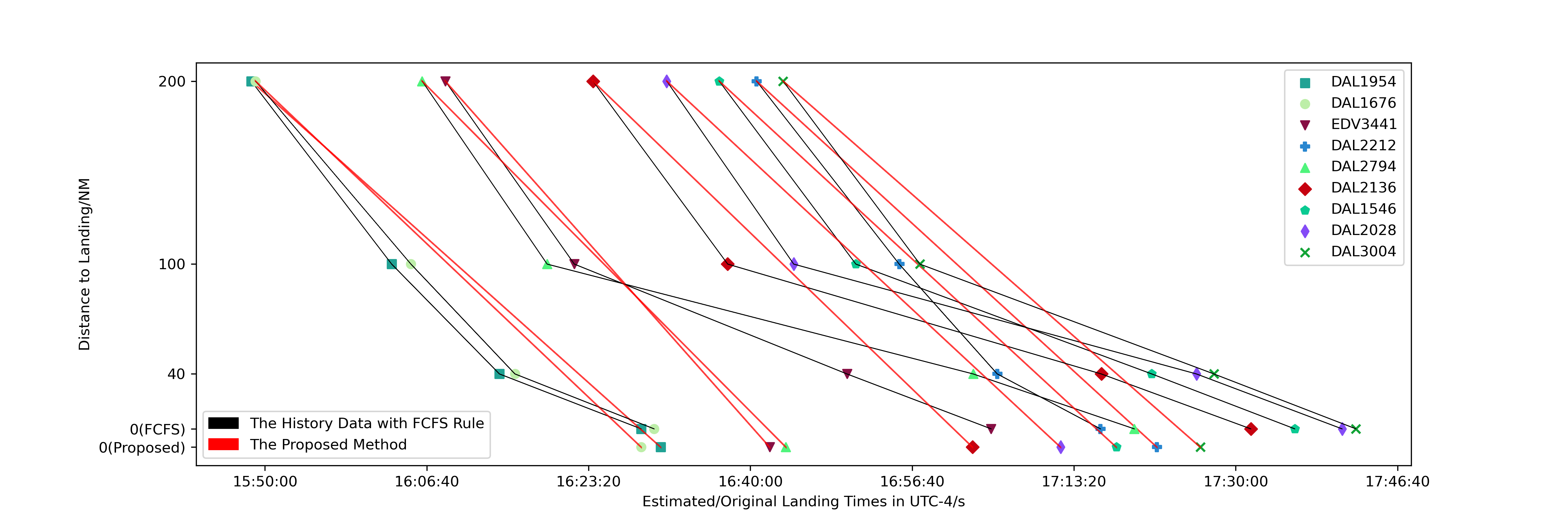}
        \caption{Case Study II: Proposed Method v.s. FCFS}
        \label{fig: results-case2}
    \end{subfigure}
    ~
    \begin{subfigure}[t]{0.95\textwidth}
        \centering
        \includegraphics[width=\textwidth]{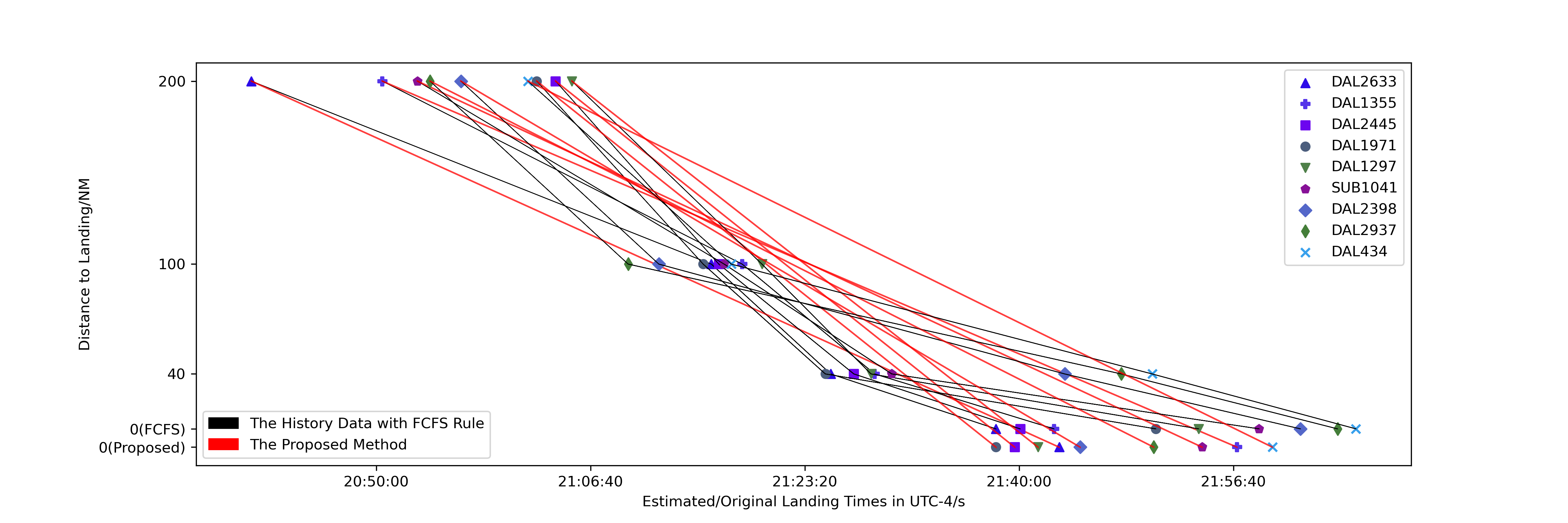}
        \caption{Case Study III: Proposed Method v.s. FCFS}
        \label{fig: results-case3}
    \end{subfigure}
    \caption{Experiment results for three case studies with landing flight: a) 13:20 to 13:45, Aug 1st, 2019; b) 16:00 to 16:30, Aug 1st, 2019; c) 21:00 to 21:20, Aug 1st, 2019. It's obvious that the proposed method requires a shorter time than the actual landing time recorded in real-world data, where FCFS rules are applied to ATC. The average landing time reduction across three case studies is 17.2\%}
    \label{fig: results-cases}
\end{figure}

We list the detailed landing scheduling results in \Cref{table: case-study-results} and \Cref{fig: results-cases}. \Cref{table: case-study-results} gives the original sequence of landing aircraft entering the $100$ NM TMA, and their predicted corresponding landing sequence, and estimated landing time from the proposed method. The MSTs between two successive landing aircraft comply with FAA regulations while incorporating additional uncertainties learned from aviation history recordings. \Cref{fig: results-cases} compares the total landing time between the FCFS rule (history recordings) and the proposed methods. As mentioned in \Cref{sec: methodologies}, the straightforward objective is to achieve a shorter total landing time for all the landing aircraft heading the same runway. In \Cref{fig: results-cases}, we color the optimized landing time in red, and the total landing time for all the aircraft is the landing time difference between the first landed aircraft and the last landed aircraft (i.e., in case I, the time duration between 13:44:47 for SKW3742 and 13:57:17 for DAL2526 is 12 minutes and 30 seconds). The average landing time reduction reaches $17.2\%$ across three case studies. The reasons for improved performance are two folds, (a) The proposed method takes advantage of the prediction power of machine learning tools by estimating the landing time distributions for each landing aircraft from historical patterns, based on several related factors (number of flight events ahead, aircraft performance factors like speed). The predicted landing time further calculates the minimum separation time of the leading aircraft, which calibrates the sub-optimal landing buffer time using the \textit{reliability} concept developed in \cite{balakrishnan2010algorithms}. (b) By incorporating the calibrated minimum separation time into the Optimization constraints, and time-constrained traveling salesman problem, the proposed method is able to find the optimal landing sequence that minimizes the total time required while satisfying the landing time interval for each aircraft. We notice that in our case studies, the number of shifts for the landing aircraft is up to 6, where a higher shifts number leads to a shorter total landing time \citep{lee2008tradeoff}

Additionally, the proposed method requires the speed up in velocity profiles for several aircraft to meet the landing sequence, which leads to the discussion on aviation sustainability.

Although aircraft emissions only account for a small percentage of total CO$_2$ emissions globally, they have a more significant impact on climate change due to their high-altitude release, and the associated contrails can amplify its warming potential. Innovations towards aviation sustainability come from three folds, (a) Aircraft technology advancement involves using lightweight materials, advanced aerodynamics, and alternative propulsion technologies like electric and hybrid-electric systems (e.g., GE Hybrid Engine). (b) alternative fuels include recycled fuels, blended fuels, or even zero-emission hydrogen fuels (e.g., sustainable fuels). (c) improve ATM efficiency with operational optimization can lead to more direct fuel consumption reduction. Following (c), various research works are proposed to include fuel consumption factors in the aircraft landing scheduling process. 

\cite{neuman1991analysis} discover that if an aircraft is allowed to speed up in TMA, and land before the earliest landing time, there will potentially be significant landing time reductions to the following aircraft. However, this obviously leads to additional fuel consumption. H. Lee \cite{lee2008fuel, lee2008study, lee2008tradeoff} investigated the tradeoff between landing scheduling algorithms and fuel consumption. Specifically, the tradeoff between speedup (time advance) and fuel consumption is investigated. Based on their aircraft landing cost model, they discover that (i) the optimization shows that allowing up to 3 minutes of time advance is optimal in most tested cases. Beyond 3 minutes, the extra fuel burn negates the savings from the reduced delay; (ii) the benefits of time advance in reducing fuel costs diminish as the number of precedence constraints (e.g. from overtaking restrictions) increases. A heavily constrained sequence leaves little flexibility to take advantage of time advance; (iii) while reducing average delay generally reduces fuel costs, the minimum fuel cost solution sometimes has higher delays than the minimum delay solution; (iv) there is no single optimal tradeoff - the balance depends on operational constraints, aircraft types, and fuel/delay costs. Overall, they provide an approach to investigate the best tradeoff given specific conditions. In recent years, there are some efforts on co-optimizing the delays and fuel costs \citep{rodriguez2019improving}. The $\varepsilon$-constraint method is shown to reduce up to $4.5\%$ total fuel consumption in a real-world case study in Madrid, Spain. However, they notice the increased computation complexity and that heavy congestion reduces opportunities for improvement, which can be the intended usage of such models.

\section{Conclusions\label{sec: conclusion}}

In this work, we propose a novel machine learning-enhanced aircraft landing scheduling algorithm, which provides a new conceptual design to avoid significant delays with safety constraints. First, the aircraft landing scheduling algorithm is formulated into a time-constrained traveling salesman problem. Being machine learning-enhanced, we incorporate machine learning-predicted results into several safety-related constraints of the time-constrained traveling salesman problem formulation. Regarding the machine learning prediction algorithm, we propose explicitly introducing nearby flight event situations and airspace complexity measurements into the conditional data-driven learner, which greatly enhances the prediction accuracy. The variable importance analysis suggests that aircraft type, ground speed, distance to destination, and airspace density are key factors affecting arrival time prediction accuracy. Finally, we evaluate and compare the performance of the proposed method through real-world case studies during peak hours at ARTCC ZTL. Various uncertainties from aircraft, speed, and airspace density are included. The key concept is to optimize the scheduling using enhanced operational predictability combining advanced instruments (e.g., ADS-B) and data analysis (e.g., arrival time prediction model in this study). 

\textbf{Insights} A few insights are discussed based on the current investigation, and a few potential research directions are suggested.

\begin{itemize}
    \item The scalability of this work can be improved. In extreme cases, the current optimization horizon corresponds to a planning horizon of $\sim10$mins. Dynamic scheduling with rolling window horizons can be integrated with the current method to extend the proposed method for a longer planning horizon. The performance of this extension can also be evaluated.
    \item The proposed study focuses on the methodology demonstration and only uses limited data at one ARTCC. A significant amount of data collection and validation at multiple airports is suggested. Model adaptiveness and generalization enhancements to multiple airports can be helpful.
    \item Weather is an important factor affecting arrival time prediction. The proposed model only considers weather on the hourly level. In future investigations, a finer-grided weather feature dataset shall be selected to validate the weather inclusion impact.
    \item Aircraft performance variables (i.e., fuel consumptions) are another group of critical factors to flight operations, as it directly impacts operating costs and environmental sustainability. We suggest including aircraft performance measures in the optimization formulation such that fuel efficiency can be directly addressed.
    \item Another important research direction is multi-runway aircraft landing scheduling, which can be especially important for ATC of major international airports. The multi-runway scheduling problem is more challenging, and significant further study is needed. Both hierarchical and concurrent optimization can be used based on our beliefs. Performance evaluation and scalability need to be balanced for decision support. 
\end{itemize}


\section*{Acknowledgment}
The research reported in this paper was supported by funds from NASA University Leadership Initiative program (Contract No. NNX17AJ86A, PI: Yongming Liu, Technical Officer: Anupa Bajwa). The support is gratefully acknowledged. 

\bibliography{ref}

\end{document}